\documentclass[journal,twoside,web]{ieeecolor}
\usepackage{tmi}
\makeatletter
\let\NAT@parse\undefined
\makeatother
\usepackage[numbers]{natbib}

\usepackage{amsmath,amssymb,amsfonts}
\usepackage{algorithmic}
\usepackage{graphicx}
\graphicspath{{figure/}}
\usepackage{textcomp}
\usepackage{bm,url}
\usepackage{multirow}
\usepackage{booktabs}
\usepackage{array}
\usepackage{makecell}
\usepackage{arydshln}
\usepackage{lettrine}
\usepackage{pifont}
\usepackage{tabularx}
\usepackage{threeparttable}
\usepackage{anyfontsize}

\usepackage[HTML]{xcolor}
\definecolor{myblue}{HTML}{008ADA}
\definecolor{table_qing}{HTML}{26b8cf}
\definecolor{table_pink}{HTML}{ff323c}
\definecolor{table_yellow}{HTML}{fdde2a}
\definecolor{mybrown}{HTML}{976431}
\usepackage[colorlinks=true, linkcolor=myblue, citecolor=myblue, urlcolor=myblue]{hyperref}
\usepackage{amsmath} 
\usepackage{xfrac}

\usepackage{xcolor}
\usepackage{etoolbox}
\usepackage{mdframed}

\usepackage{xcolor}
\usepackage{etoolbox}

\newbool{showred}
\setbool{showred}{true}

\newcommand{\tred}{{}}

\def\BibTeX{{\rm B\kern-.05em{\sc i\kern-.025em b}\kern-.08em
		T\kern-.1667em\lower.7ex\hbox{E}\kern-.125emX}}
\begin{document}
	\IEEEaftertitletext{\vspace{-10pt}}
	\title{\huge{Efficient Large-Deformation Medical Image Registration via Recurrent Dynamic Correlation}}
	
	\author{
		Tianran Li, \and Marius Staring, Yuchuan Qiao
		\thanks{This work is supported by the National Natural Science Foundation of China under Grant 82102002. Corresponding author: Yuchuan Qiao. Email: YuchuanQiao@fudan.edu.cn.}
		\thanks{Tianran Li and Yuchuan Qiao are with the Institute of Science and Technology for Brain-Inspired Intelligence, Fudan University, Shanghai, China. Marius Staring is with the Department of Radiology, Leiden University Medical Center, Leiden, The Netherlands.}
	}
	\maketitle

	\begin{abstract}
		Deformable image registration estimates voxel-wise correspondences \tred{between images through} spatial transformations, and \tred{plays a key role} in medical imaging.
		While deep learning methods have significantly reduced runtime, efficiently handling large deformations remains a challenging task. Convolutional networks aggregate local features but lack direct modeling of voxel correspondences, promoting recent works to explore explicit feature matching. Among them, voxel-to-region matching is more efficient for direct correspondence modeling by computing local correlation features whithin neighbourhoods, while region-to-region matching incurs higher redundancy due to excessive \tred{correlation pairs} across large regions. However, the inherent locality of voxel-to-region matching hinders the capture of long-range correspondences required for large deformations. 
		To address this, we propose a Recurrent Correlation-based framework that dynamically relocates the matching region toward more promising positions. At each step, local matching is performed with low cost, and the estimated offset guides the next search region, supporting efficient convergence toward large deformations. In addition, we uses a lightweight recurrent update module with memory \tred{capacity} and decouples motion-related and texture features to suppress semantic redundancy. 
		We conduct extensive experiments on brain MRI and abdominal CT datasets under two settings: with and without affine pre-registration.
		Results show that our method exibits a strong accuracy-computation trade-off, surpassing or matching the state-of-the-art performance.  
		For example, it achieves comparable performance on the non-affine OASIS dataset, while using only 9.5\% of the FLOPs and running 96\% faster than RDP, a representative high-performing method.
	\end{abstract}
	
	\vspace{-4pt}
	
	\vspace{-7pt}
	\section{Introduction}
	\label{sec:introduction}
	\vspace{0pt}
	
	\lettrine{D}{\fontsize{12}{14}\selectfont eformable} medical image registration aligns 3D image pairs (named the fixed image $I_f$ and the moving image $I_m$) by applying spatial transformations to match their structures within a common coordinate system \cite{survey}. It involves mapping each voxel in $I_f$ to a corresponding most relevant location in $I_m$. This process is crucial in clinical applications like tumor monitoring and CT/MRI/PET data fusion \cite{fusion}. 
	Traditionally, it involves iterative optimization to find the optimal deformation parameters with extensive and time-consuming searches \cite{survey}. 
	\tred{
		Although GPU acceleration has made iterative registration much faster, these methods are still sensitive to starting conditions during optimization.
	}
	
	\tred{
		In contrast, deep learning methods could complete registration in seconds by estimating the deformation field from scratch \cite{learningsurvey}.
	}
	However, efficiently handling large deformations remains challenging.
	U-shaped convolutional structures which directly predict the deformation field in an end-to-end manner \cite{voxelmorph,midir}, struggle with large deformations due to the limited receptive fields. To mitigate this limitation, pyramid \cite{lapirn,pcreg}, recursive \cite{vit} and hybrid architectures \cite{sdhnet, rdp} are later introduced, which decompose large deformations into smaller and more manageable steps. Although they have been proven to be effective to some extent for large deformations, the convolution operations are fundamentally the weighted aggregation within neighborhoods, which is less capable to model voxel-wise correspondences between the two images.
	
	Several recent works introduce explicit feature matching by establishing voxel-wise dense matches within a specified region, a process that naturally aligns with the core objective of the registration task: identifying voxel-level correspondences between fixed and moving images. 
	In this sense, such process can be viewed as {\itshape a search within specific regions to identify the most relevant target coordinates}.
	However, existing methods still struggle to balance accuracy and computational efficiency.

	\tred{
	One group of these methods \cite{6dcorr, transmorph, xmorpher, transmatch, cgnet} adopts a {\itshape \textbf{region-to-region}} matching strategy, typically implemented via the attention mechanism in Transformer \cite{vit, swin}, which densely computes pairwise voxel correlations within large regions of image pair.
	}
	However, many of these voxel pairs are irrelevant due to the high structural similarity in medical images \cite{voco}, resulting in substantial computational redundancy.
	In contrast, another line of work \cite{dualprnetplus, corrmlp} adopts {\itshape \textbf{voxel-to-region}} matching, where each voxel in $I_f$ independently queries its corresponding regions in $I_m$. Compared to {\itshape region-to-region} strategies that correlate all voxels within a window, {\itshape voxel-to-region} matching assigns each voxel a specific search scope, reducing unnecessary pairwise correlation operations. Typically, to keep the computational cost manageable, the region in $I_m$ is confined to a small local neighborhood, which is often insufficient to capture very large deformations. CorrMLP \cite{corrmlp} attempts to mitigate this by applying multiple large-window MLPs, but at a cost of significantly increased computation.
	In addition, correlation features are often treated as auxiliary channels concatenated with the two image features \cite{dualprnetplus, corrmlp}, which introduces additional semantic redundancy and reduces their effectiveness in guiding correspondence search.

	\begin{figure}[!t]
		\vspace{-1pt}
		\centerline{\includegraphics[width=0.485\textwidth]{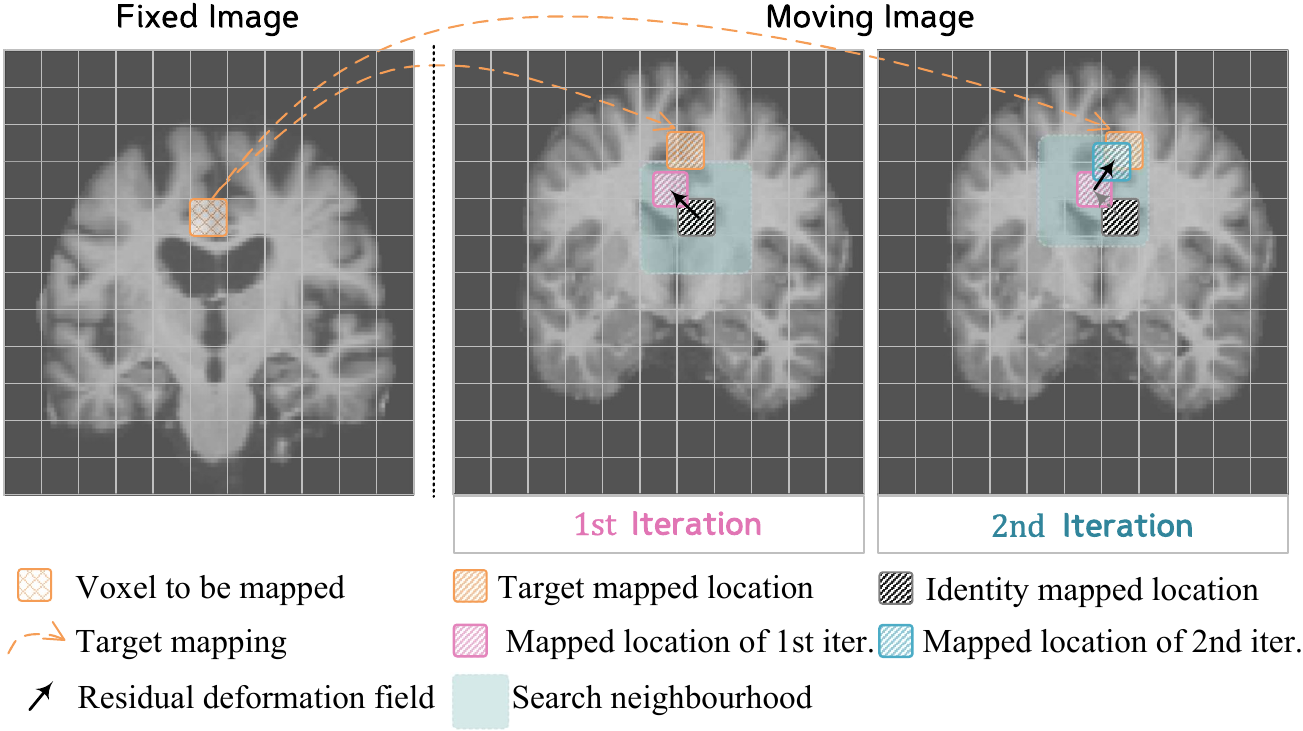}}
		\vspace{-3pt}
		\caption{\fontsize{7.9}{9.4}\selectfont\textit{Illustration of the recurrent local search strategy of ReCorr.} A voxel from the fixed image (left) is progressively matched to its corresponding location in the moving image via \tred{recurrent} local search and \tred{dynamic updates}. At each iteration, the search is performed within a local neighborhood (blue), and the deformation field is incrementally updated (black arrows), guiding the search region toward more accurate correspondences.}
		\label{fig:notion}
		\vspace{-12pt}
	\end{figure}
	\renewcommand{\baselinestretch}{1}

	In this work, we rethink how large deformations are modeled, and explore to efficiently establish long-range voxel correspondences. To overcome the locality of the {\itshape voxel-to-region} strategy while still leveraging its computational efficiency, we introduce an efficient Recurrent Correlation-based framework, named ReCorr, which performs dynamic local search guided by iterative search-center relocation. The basic idea is illustrated in Fig. \ref{fig:notion}. At each step, the model conducts {\itshape voxel-to-region} matching within a local neighborhood, then dynamically updates the search center to guide the next iteration. Through this recurrent process, the search windows adaptively shift toward more promising matching regions and gradually converge to the global optimal correspondence, with each step at a low computational cost. This search scheme is inspired by gradient-based optimization, where parameters are incrementally refined toward convergence.

	Besides, we design a lightweight recurrent update module \tred{with memory capacity that retains useful deformation context} across iterations. To help the model focus on spatial alignment without \tred{interference from redundant semantic features}, instead of directly concatenating all features, we decouple the prediction into two branches: one for the motion-related information and the other for the image texture. Our module operates at a single resolution with shared parameters, reducing the inference time compared to previous iterative methods \cite{vit, sdhnet} which jointly use multi-scale image features. ReCorr adopts a pyramid architecture, where iterative search begins at a coarse resolution to provide a reliable initialization for finer levels. Unlike traditional coarse-to-fine schemes, the iterations are not limited to the number of pyramid levels, which effectively reduces low-resolution errors by allowing continuous refinement at each scale.

	To evaluate our approach for both small and large deformations, we conducted experiments in two scenarios: \textit{regular} experiments on datasets with affine pre-registration, and \textit{extreme} trials on datasets without such pre-registration, formulated to present more complex and challenging deformations. Across all experimental setups on two brain MRI datasets and an abdominal CT dataset, our method consistently outperforms or compares favorably with the state-of-the-art methods but at significantly lower FLOPs and inference time.

	\begin{figure*}[!ht]
		\vspace{-4pt}
		\centerline{\includegraphics[width=0.985\textwidth]{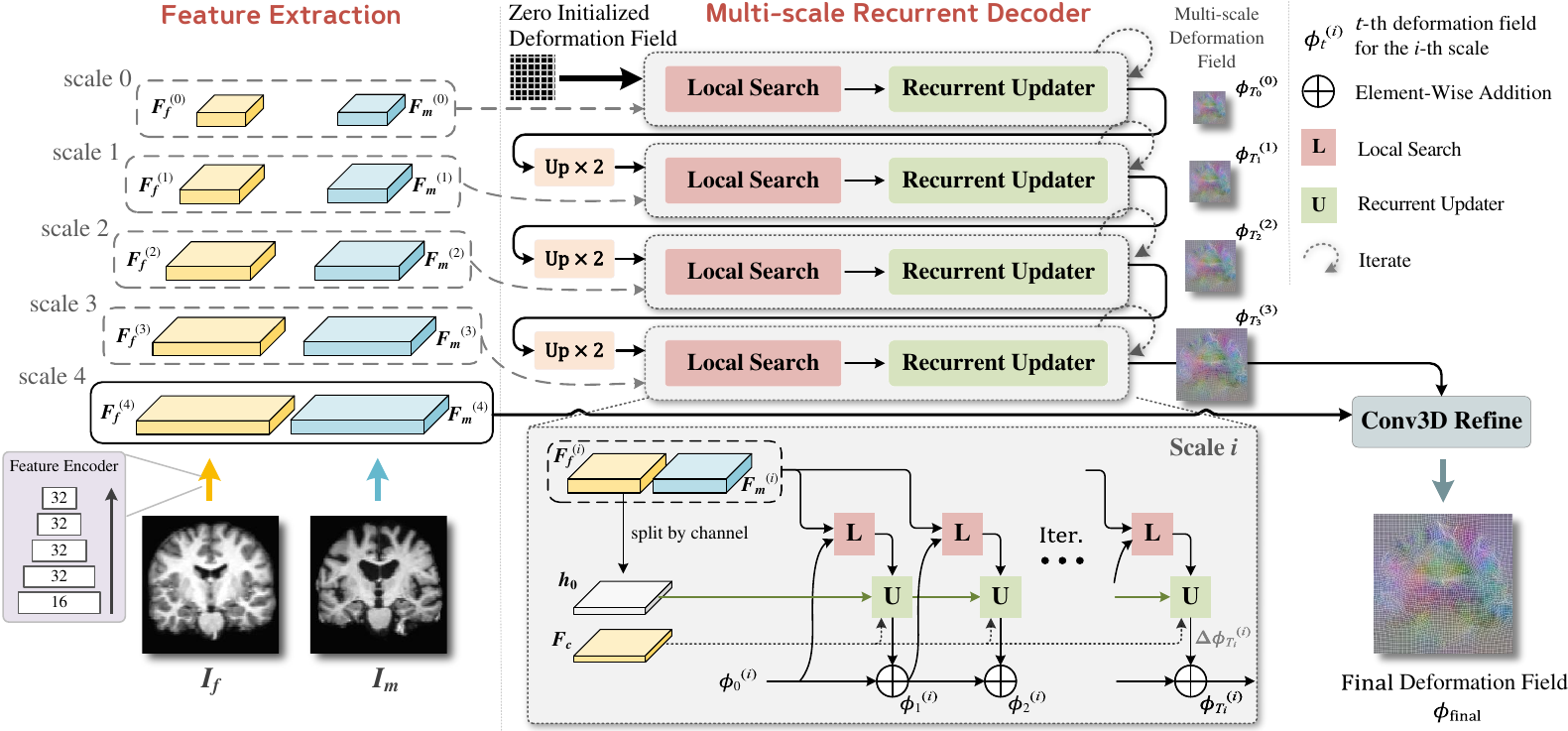}}
		\caption{\fontsize{7.9}{9.5}\selectfont Architecture overview of the proposed ReCorr. Given a pair of images, ReCorr first extracts multi-scale features using a same encoder, then performs iterative refinement from scale $0$ to scale $3$. At each scale $i$, the $t$-th iteration begins by conducting local search and then updates the residual deformation field $\Delta \phi_t^{(i)}$, which is added to the deformation field from the previous iteration. Except for the zero-initialized scale 0, each scale is initialized by upsampling the final output from the preceding scale. The final deformation field $\phi_{\rm final}$ is refined at the original resolution using a Conv3D block.}
		\label{fig_overview}
		\vspace{-4pt}
	\end{figure*}
	
	In summary, this work has the following contributions:
	\vspace{0pt}
	
	\subsubsection{Efficient Search Scheme for Large Deformations} We propose ReCorr with an efficient search scheme for large deformations by dynamically performing voxel-to-region matching with iterative search-center relocation. The pyramid recurrent process compensates for the spatial limitation of voxel-to-region matching, enabling progressive convergence toward the global optimum at a low computational cost per step.

	\subsubsection{Lightweight Recurrent Update Module with Motion-texture Decoupling} 
	A lightweight update module is introduced to retain historical deformation cues via memory across iterations. To suppress semantic redundancy and focus on alignment, we decouple the prediction into two branches for motion-related information and image texture.

	\subsubsection{Excellent Accuracy-computation Trade-off on Diverse Deformation scenarios} 
	We conduct extensive evaluations on both regular (affine pre-registered) and extreme (unregistered) settings across two brain MRI datasets and one abdominal CT dataset. Our method consistently achieves superior or competitive accuracy compared to state-of-the-art approaches, while significantly reducing FLOPs and inference time.

	\vspace{2pt}
	We \tred{review} related work in Section~\ref{sec:Related Work}, and \tred{presents our method} in Section~\ref{sec:Method}, \tred{including the problem definition, overall architecture, local search module, and recurrent updater}. Section~\ref{sec:experiment} describes the experimental setup, and Section~\ref{sec:results} reports the results and analysis. Finally, discussions and conclusions \tred{are provided} in Section~\ref{sec:conclusion}. \tred{Our code is available at {\itshape https://github.com/YQiaoGroup/Registration-ReCorr}.}
	
	\vspace{4pt}
	\section{Related Work}
	\label{sec:Related Work} 
	\vspace{2pt}
	\subsection{Non-learning Based Medical Image Registration}
	\vspace{-1pt}

	Image registration has traditionally been approached as an optimization problem, balancing between a similarity term and a regularization term. In the early stages, the methods include elastic registration methods \cite{elastic}, B-spline based free-form deformations \cite{bspline}, and the Demons algorithm \cite{demons}. Later, there are several techniques focusing on diffeomorphic transformations for anatomical accuracy, such as the diffeomorphic Demons \cite{diffdemons}, Large Deformation Diffeomorphic Metric Mapping (LDDMM) \cite{lddmm} and Symmetric Normalization (SyN) \cite{SyN}. These methods, while effective and robust, tend to be time-intensive because of the time-consuming and resource-intensive optimization. \tred{With the adoption of GPU acceleration, many iterative methods have become faster, but they still remain sensitive to initialization and prone to local optima.
	}

	\vspace{-10pt}
	\subsection{Learning Based Medical Image Registration}
	
	\textbf{\textit{Pure Convolutional Networks.  }}
	Recently, learning-based methods have demonstrated the potential for fast and high-precision image registration \cite{nonrigid2017, de2019deep}.
	\tred{
		A milestone is VoxelMorph \cite{voxelmorph}, which introduced an unsupervised U-shaped convolutional framework for predicting deformation fields from image pairs, inspiring numerous subsequent approaches.
	}
	MIDIR \cite{midir} use a similar structure to VoxelMorph but modifies the output to be the parameters of the B-spline transformation.
	\tred{These methods build on pure convolutional operations and} predict dense deformation fields \tred{in a single pass}. While some other work attempt to improve accuracy by enhancing feature representations \cite{robust, dino, fewshot} or reducing redundant parameters \cite{decoderonly}, they still face challenges with large deformations due to the inherently limited receptive fields.
	To tackle large deformations, some methods adopt progressive refinement via recursive or pyramid structures. VTN \cite{vtn} stacks multiple U-Net-based subnetworks, while LapIRN \cite{lapirn} introduces a coarse-to-fine pyramid strategy that has become widely adopted. 
	Other variants of pyramid-based methods include \cite{pcreg, dualprnet, pacs, mlvirnet}.
	Several works combine the pyramid structure with recurrent refinement. SDHNet \cite{sdhnet} jointly estimates and fuses multi-scale deformation fields at each iteration, but its parallel design increases structural complexity and inference latency due to the need for synchronization across levels. In contrast, RDP \cite{rdp} applies recurrent loops sequentially across pyramid levels, yet suffers from significant computational overhead due to repeated feature concatenation and heavy 3D convolutions. IIRP \cite{iirp} follows a similar strategy but introduces additional cost by performing NCC-based early stopping at each iteration. However, these methods are pure convolutinal architectures, which are insufficient to capture voxel-correspondences. 
	
	\vspace{0pt}
	\textbf{\textit{Region-to-region Explicit Feature Matching. }}
	Recent works introduce explicit voxel-wise matching to establish dense correspondences. A common strategy is region-to-region matching, where pairwise voxel correlations are densely computed between two predefined regions. The pioneering work \cite{6dcorr} introduces a 6D correlation feature computed over the whole image pair. Later, the rise and popularity of Transformer architectures have led to methods such as \cite{transmorph, xmorpher,transmatch}, which adopt Transformer-based architectures for voxel-wise correlation modeling, typically within partitioned windows of the input images. Inspired by these methods, CGNet \cite{cgnet} introduces a modified correlation module as an alternative to the attention mechanism in Transformers, enabling efficient processing of high-resolution features.
	However, due to the relatively consistent contextual positions of structures in medical images \cite{voco}, many voxel pairs within the window are irrelevant, leading to significant computational redundancy.

	\vspace{0pt}
	\textbf{\textit{Voxel-to-region Explicit Feature Matching. }}
	Several other methods adopt a more efficient voxel-to-region strategy, where each voxel queries a local neighborhood in the other image to identify the best match, reducing redundancy by avoiding exhaustive pairwise voxel correlations. However, their locality limits the ability to capture large deformations.
	DualPRNet++ \cite{dualprnet} incorporates the local correlation features as auxiliary information in a dual-stream pyramid network to enhance deformable image registration. 
	CorrMLP \cite{corrmlp} combines the correlation features with multi-window MLPs to capture a broader receptive field, but the fully-connected nature of MLPs leads to considerable computational overhead, especially at high resolutions.
	Notably, these methods treat correlation features merely as supplementary cues, simply concatenating them with image features within a feature-based framework. In contrast, we highlight the role of local correlation in explicitly modeling voxel-wise spatial relationships and decouple the update process into two branches: one for motion-related information and the other for image texture, to better focus on alignment and reduce semantic redundancy.
	
	While our method is connected to the existing methods, our motivation is originated from the efficiency–accuracy trade-off challenges for large-deformation registration and differs substantially in design. 
	We explicitly leverage the voxel-to-region matching paradigm for its direct role in establishing voxel correspondences, and address its locality through pyramid recurrent refinement. This enable the model to perform dynamic local search, with search centers progressively shifting toward more promising regions, achieving accurate long-range matching at low computational cost.
	In addition, by decoupling prediction into motion and texture branches, our model focuses on spatial alignment without being distracted by semantic redundancy, improving both precision and efficiency.

	\vspace{6pt}
	\section{Method}
	\label{sec:Method} 
	\vspace{4pt}

	\subsection{Definition}
	\label{sec:Method:definition}
	\vspace{-2pt}
	
	Given a moving image $I_m \in \mathbb R^3$ and a fixed image $I_f \in \mathbb R^3$, deformable image registration seeks a dense, non-linear transformation $\phi: \mathbb R^3 \rightarrow \mathbb R^3$ from $I_m$ to $I_f$, such that the deformed image $I’_m = I_m \circ \phi$ is as similar as possible to $I_f$. The deformation field $\phi$ defines voxel-wise correspondences by mapping each voxel in $I_f$ to a corresponding location in $I_m$. In this context, handling large deformations means establishing long-range voxel correspondences across the two images.
	\vspace{-3pt}
	\subsection{Architecture of ReCorr}
	\label{sec:Method:overview}
	\vspace{-0pt}
	To efficiently handle large deformations in medical image registration, 
	we propose a recurrent correlation-based framework specifically designed to establish long-range voxel correspondences with low computational cost.
	Our method leverages the voxel-to-region matching paradigm and decomposes the large distance into several shorter steps through a pyramid recurrent refinement strategy. At each level, the model iteratively performs dynamic local search by matching voxels in $I_f$ to local neighborhoods in $I_m$, with search centers progressively updated toward more likely correspondences, thus effectively bridging the spatial gap.
	An overview of the proposed framework is given in Fig.~\ref{fig_overview}.
	
	Given a pair of input images $I_f$ and $I_m$, we use a light-weight encoder akin to that in VoxelMorph~\cite{voxelmorph} to extract multi-scale features, as shown in Fig.~\ref{fig_overview} {\itshape (left)}. This encoder is shared by the two images and generates two sets of features at scales of \scalebox{1.02}{$\sfrac{1}{16}$}, \scalebox{1.02}{$\sfrac{1}{8}$}, \scalebox{1.02}{$\sfrac{1}{4}$}, \scalebox{1.02}{$\sfrac{1}{2}$}, and the original resolution, represented as $\{{\boldsymbol F_f^{(i)}}\}_{i=0,1,2,3,4}$ and $\{{\boldsymbol F_m^{(i)}}\}_{i=0,1,2,3,4}$, respectively. 
	Benefiting from the inductive bias of CNNs in translation equivariance and locality, these features capture local structures and texture patterns, which are suitable for voxel-wise matching.
	Features at \scalebox{1.02}{$\sfrac{1}{16}$}, \scalebox{1.05}{$\sfrac{1}{8}$}, \scalebox{1.05}{$\sfrac{1}{4}$} and \scalebox{1.02}{$\sfrac{1}{2}$} resolutions ($\{{\boldsymbol F_f^{(i)}}\}_{i=0,1,2,3}$) are used for iterative updates, while features at original resolution (${\boldsymbol F_m^{(4)}}$) are used for final refinement. 
	A portion of the features from the fixed image, denoted as $\boldsymbol F_c$, is used to preserve image texture and serves as context guidance during deformation estimation (see Section \ref{sec:Method:update} for details).
	Note that the feature extraction process is executed only once and the extracted features are consistently reused across all iterations for efficiency.

	Starting from scale 0, ReCorr performs iterative updates sequentially up to scale 3. The procedures at all the four scales are the same, and we take scale $i$ ($i=0,1,2,3$) as an example. At scale $i$, there are $T_i$ iterations. During the $t$-th iteration, we perform local search by querying each voxel in $I_f$ within a local neighborhood centered at its previously estimated mapped location in $I_m$, resulting a dynamic 4D correlation feature $\mathcal C$.
	Subsequently, $\mathcal C$ is fed into a lightweight recurrent updater to produce the residual deformation field $\Delta\phi^{(i)}_t$, and the current deformation field $\phi^{(i)}_t$ is updated as
	\begin{equation}
		\phi^{(i)}_t = \phi^{(i)}_{t-1} + \Delta\phi^{(i)}_t, \ t=1,2,...,T_i.
	\end{equation}
	\vspace{-2pt}
	As a result, the final deformation field of the $i$-th scale $\phi^{(i)}_{T_i}$ is the accumulation of all learned $\Delta\phi^{(i)}$:
	\vspace{-2pt}
	\begin{equation}
		\phi^{(i)}_{T_i} = \phi^{(i)}_0 + \sum_{t=1}^{T_i}\Delta\phi^{(i)}_t, \ t=1,2,...,T_i, 
		\vspace{-5pt}
	\end{equation}
	where $\phi^{(i)}_0$ denotes the initial deformation field at scale $i$. At scale 0, $\phi^{(0)}_0$ is zero-initialized for identical mapping, while at higher scales, $\phi^{(i)}_0$ is derived by upsampling the final deformation field from the preceding scale with a factor of 2.
	Finally, after all iterations are completed, the deformation field is upsampled to full image resolution and refined using the high-resolution ${\boldsymbol F_f^{(4)}}$ and ${\boldsymbol F_m^{(4)}}$ to recover fine-grained details, yielding in the final deformation field $\phi_{\rm final}$.

	\vspace{-3pt}
	\tred{
		\subsection{Local Search Module}
	}
	\label{sec:Method:correlation}
	\vspace{-0pt}
	\begin{figure}[!htb]
		\centerline{\includegraphics[width=0.49\textwidth]{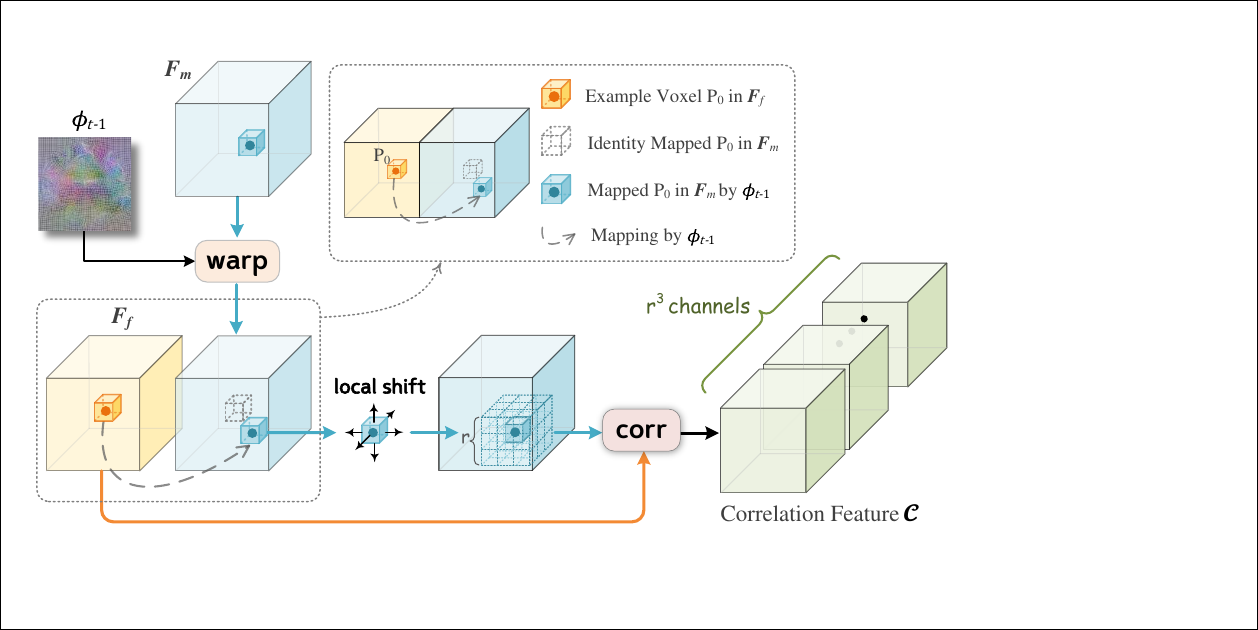}}
		\caption{The structure of the local search module. Note that we omit the scale index `$i$' for simplicity.}
		\label{fig_localsearch}
		\vspace{-3pt}
	\end{figure}

	Motivated by \cite{dualprnetplus}, ReCorr uses voxel-to-region local correlation to explicitly model voxel correspondences, in a manner that closely aligning with the goal of registration. This operation is referred to as {\itshape local search}, as each voxel in $I_f$ queries within a restricted neighborhood in $I_m$ for the best match.	
	It takes as input the extracted features ${\boldsymbol F_f^{(i)}}$ and ${\boldsymbol F_m^{(i)}}$ ($i$=0,1,2,3), along with the previous deformation field $\phi_{t-1}^{(i)}$, to dynamically construct a lightweight 4D correlation feature. For simplicity, the scale index '$i$' is omitted in the following descriptions. Both ${\boldsymbol F_f}$ and ${\boldsymbol F_m}$ are assumed to have the shape $B\times C\times D\times H \times W$, where $B$ is the batch size, $C$ is the number of channels, and $D$, $H$, $W$ denote the spatial dimensions.
	
	As illustrated in Fig.~\ref{fig_localsearch}, the computation of local search during the $t$-th iteration involves three steps:\\
	1) Global Localization: $\boldsymbol F_m$ is warped using $\phi_{t-1}$ to obtain the roughly mapped global locations.\\
	2) Local Shift: Centered around the identified mapped locations, cube-shaped local neighbourhoods with a side $r$ voxels are extracted by applying padding and spatial shifting. Specifically, the warped feature ${\boldsymbol F_m}$ is padded and then shifted along the $x$, $y$ and $z$ axes within the range $\left[-r//2, r//2\right]$, producing a total of $r^3$ offsets positions.\\
	3) Correlation and Aggregation: For each shift position, a correlation volume $\boldsymbol {Corr}$ is computed between $\boldsymbol F_f$ and the corresponding shifted, warped $\boldsymbol F_m$ using the normalized dot product of feature vectors:
	\begin{equation}
		{\boldsymbol {Corr}}_{bzyx} = \frac{1}{C} \left\langle \left(\boldsymbol F_f\right)_{b\cdot zyx}, \ \mathbf{Shift}(\boldsymbol F_m \circ \phi_{t-1})_{b\cdot zyx} \right\rangle,
	\end{equation}
	
	where $b,z,y,x$ index the batch and spatial dimensions $B$, $D$, $H$, $W$, respectively. The resulting $\boldsymbol {Corr}$ has shape $B\times D\times H\times W$. All correlation volumes across the $r^3$ offsets are then concatenated along the channel dimension to form a 4D correlation feature $\boldsymbol{\mathcal C} \in B\times r^3\times D\times H\times W$. The hyper-parameter $r$ is set to 3 based on ablation studies. 
	A local neighbourhood with side length $r$ at coarse scales corresponds to a much larger region in the original resolution space, enabling broader and more efficient search. In contrast, higher-scales focus the search within smaller regions, allowing more precise voxel-level matching and refinement.

	The computational complexity of local correlation is $\mathcal{O}(C\cdot r^3\cdot DHW)$. Compared to convolutional operations, which implicitly aggregate local information with complexity $\mathcal{O}(C\cdot C_{out}\cdot k^3\cdot DHW)$ ($C$, and $C_{out}$ are the input and output channel dimensions), local correlation offers both direct correspondence modeling and lower cost. Compared to region-to-region matching, which also performs explicit feature matching but requires dense correlations between all voxel pairs across two regions, resulting in much higher cost at $\mathcal{O}(C\cdot (DHW)^2)$ if applied to the full volume \cite{vit}. Even window-based variants \cite{swin, transmorph} that limit matching to partitioned subregions still incur substantial overhead, particularly in 3D where the number of windows grows cubically with resolution. 
	
	Overall, local correlation is both task-aligned and computationally efficient, but its performance is limited by the narrow search range. To address this, we introduce a pyramid recurrent strategy that gradually guides the search toward more accutate correspondences, while maintaining low computational cost.

	\vspace{-4pt}
	\subsection{Recurrent Updater}
	\label{sec:Method:update}
	\vspace{-6pt}
	\begin{figure}[!ht]
		\centerline{\includegraphics[width=0.49\textwidth]{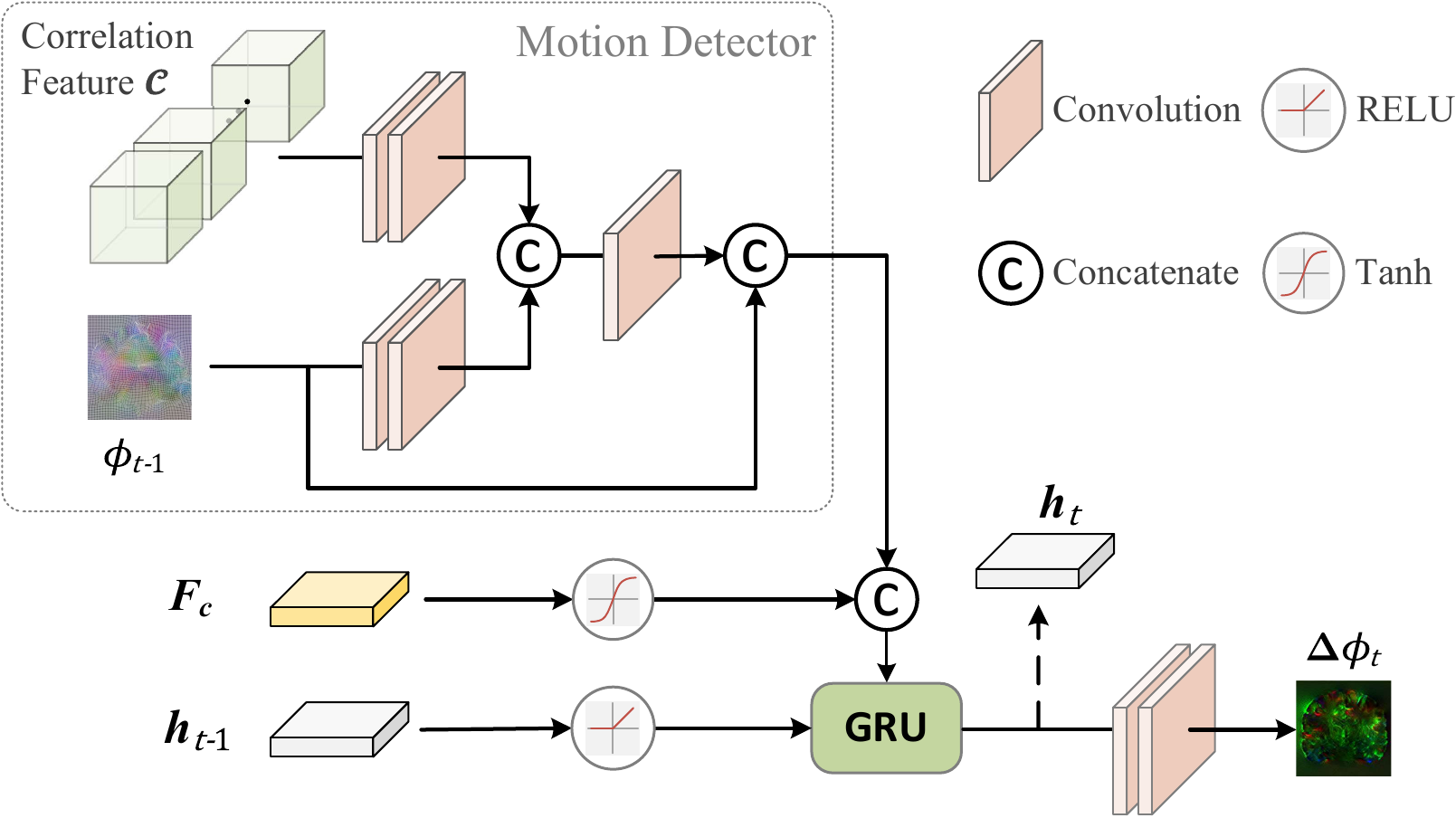}}
		\vspace{-5pt}
		\caption{The structure of the recurrent updater for each scale.}
		\label{fig_update}
		\vspace{-8pt}
	\end{figure}
	
	We propose a lightweight recurrent updater to iteratively update the deformation field, which in turn guides the next iteration. To focus on spatial alignment, the prediction is decoupled into motion and texture branches.
	
	At the $t$-th iteration of scale $i$, the updater predicts a residual deformation field $\Delta\phi^{(i)}_t$ based on the previous estimate $\phi^{(i)}_{t-1}$. All iterations at the same scale share the same update module. For clarity, we omit the scale index '$i$' in the following.  
	
	As show in Fig. \ref{fig_update}, the core of the updater is a gated recurrent unit (GRU) \cite{gru}, which maintains a hidden state to propagate deformation-related cues across iterations. The correlation feature $\boldsymbol{\mathcal C}$ and the previous deformation field $\phi_{t-1}$, which both reflect spatial relationships between the fixed and moving images, are treated as motion-related inputs and processed by a motion detector to produce $\boldsymbol m_t$.
	
	Meanwhile, the fixed feature map $\boldsymbol F_f$ is split along the channel dimension: one half initializes the GRU hidden state $\boldsymbol h_0$, and the other is used as the contextual guidance $\boldsymbol F_c$ to preserve image texture. Then, the motion feature $\boldsymbol m_t$, the activated context $\boldsymbol F_c$, and the previous hidden state $\boldsymbol h_{t-1}$ are concatenated and fed into the GRU. The updated hidden state $\boldsymbol h_t$ is computed as follows:
	\vspace{-1pt}
	\begin{equation}
		\vspace{-2pt}
		\begin{aligned}
			\boldsymbol {z_t} & =\mathrm{\sigma}\big(\mathrm{Conv}(\left[\boldsymbol h_{t-1}, \boldsymbol m_t, \boldsymbol F_c\right], \boldsymbol W_z)\big), \\
			\boldsymbol r_t & =\mathrm{\sigma}\big(\mathrm{Conv}([\boldsymbol {h_{t-1}}, \boldsymbol m_t, \boldsymbol F_c], \boldsymbol W_r)\big), \\
			\tilde{\boldsymbol h}_t & =\mathrm\tanh \big(\mathrm{Conv}([\boldsymbol r_t \odot \boldsymbol h_{t-1}, \boldsymbol m_t, \boldsymbol F_c], \boldsymbol W_h)\big), \\
			\boldsymbol h_t & =(1-\boldsymbol z_t) \odot \boldsymbol h_{t-1}+\boldsymbol z_t \odot \tilde{\boldsymbol h}_t.
		\end{aligned}
	\end{equation}
	Inspired by \cite{inception2}, the convolution operators within the GRU unit are designed to be separatable, implemented as sequential GRU branches with $1\times 1\times 5$, $1\times 5\times 1$, and $5\times 1\times 1$ convolutions. This design enlarges the receptive field without significantly increasing the model complexity. Finally, the updated hidden state $\boldsymbol h_t$ is passed through a convolution layer to produce the residual deformation field $\Delta\phi_t$.

	\begin{figure*}[!ht]
		\centering
		\begin{minipage}[t]{0.48\textwidth}
			\centering
			\includegraphics[width=\linewidth]{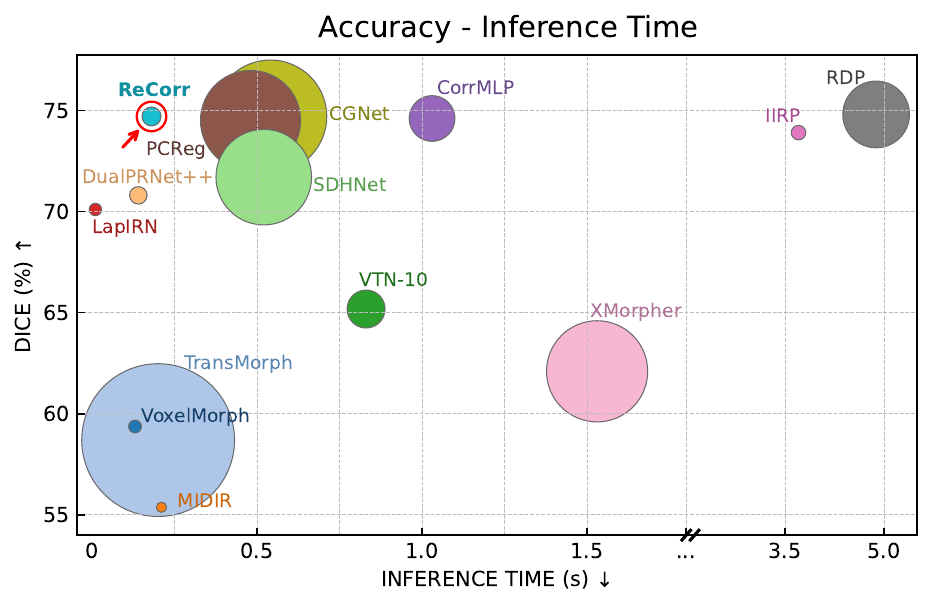}
		\end{minipage}
		\hfill
		\begin{minipage}[t]{0.489\textwidth}
			\centering
			\includegraphics[width=\linewidth]{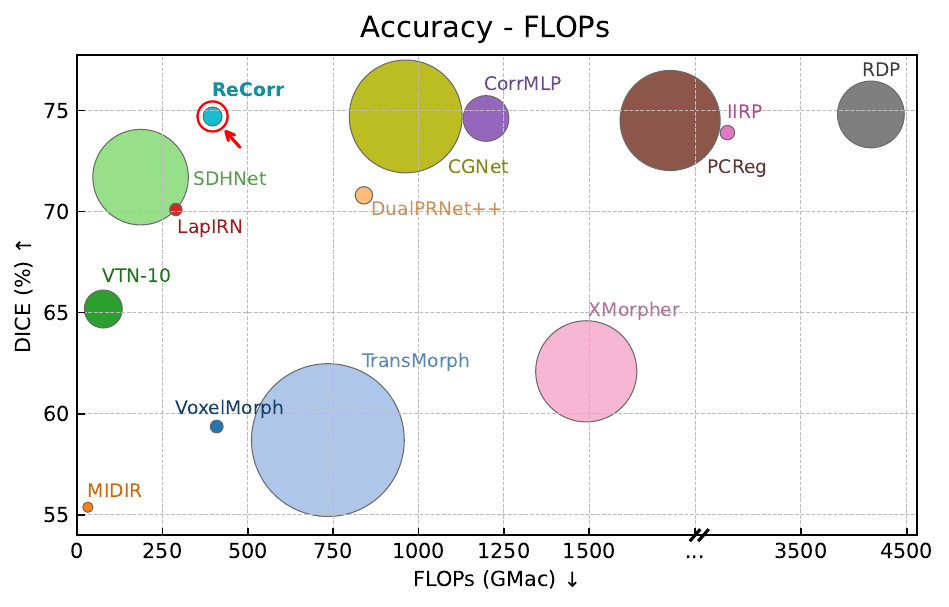}
		\end{minipage}
		\caption{The two scatter plots compare registration accuracy (Dice) against inference time (left), and FLOPs (right) on the OASIS dataset without affine pre-registration. Circle size represents model parameters, where a \textbf{larger} circle indicates \textbf{more} parameters. Arrows ($\uparrow\!\!\! \ \downarrow$) indicate the preferable direction of metrics. The proposed ReCorr achieves a favorable trade-off between accuracy and efficiency in both views.}
		\label{fig:trade-off}
		\vspace{-3pt}
	\end{figure*}

	\vspace{-6pt}
	\subsection{Unsupervised Loss Function}
	\vspace{-1pt}
	
	ReCorr produces $T_i$ deformation fields $\{\phi_1^{(i)}, \phi_2^{(i)}, \dots, \phi_{T_i}^{(i)}\}$ at each scale $i$ $(i=0,1,2,3)$. Togather with the full-resolution refined result $\phi_{\rm final}$, the complete sequence of predicted deformation fields is
	\begin{equation*}
		\{\{\phi_1^{(0)}, \dots, \phi_{T_0}^{(0)}\}, \cdots, \{\phi_1^{(3)},\dots,\phi_{T_3}^{(3)}\}, \phi_{\rm{final}}\}.
	\end{equation*}
	For loss computation, all deformation fields are upsampled to the original resolution, yielding a unified sequence $\{\phi_1, \phi_2, \dots, \phi_t,\dots,\phi_{T}\}$, where $T = T_0 + T_1 + T_2 + T_3 + 1$.
	Each $\phi_t$ is supervised using two terms: an image similarity loss and a deformation field regularization loss, following \cite{voxelmorph}. The combined loss function is denoted as $\mathcal L_{\text{single}}$, defined as:
	\begin{equation}
		\mathcal{L}_{\text{single}}(I_f, I_m, \phi_t) = \mathcal{L}_{\text{sim}}(I_f, I_m \circ \phi_t) + \lambda \mathcal{L}_{\text{reg}}(\phi_t),
	\end{equation}
	where $\mathcal{L}_{\text{sim}}$ denotes the similarity loss, implemented using either mean squared intensity error (MSE) or normalized cross-correlation (NCC), and $\mathcal{L}_{\text{reg}}$ is an L2 penalty on the spatial gradients of $\phi_t$. The hyperparameter $\lambda$ balances the trade-off of two terms .
	
	To ensure effective supervision across multiple outputs, we apply exponentially increasing weights to the sequence $\{\phi_1, \phi_2, \dots, \phi_t,\dots,\phi_{T}\}$, following \cite{raft}:
	\vspace{-2pt}
	\begin{equation}
		\mathcal{L} = \sum_{t=1}^{T} \gamma^{T-t} \mathcal{L}_{\text{single}}(I_f, I_m, \phi_t),
		\vspace{-2pt}
	\end{equation}
	where $\gamma\in(0,1]$ and is set to $0.7$ in our experiments. 
	This strategy assigns lower weights to early predictions and progressively increases the weights for later ones, ensuring greater emphasis on the later stages of deformation field prediction, which contribute more to the final result.
	\vspace{-5pt}
	\subsection{Differomorph Version of ReCorr}
	\vspace{-2pt}

	In addition to the standard version, we offer a diffeomorphic variant of ReCorr, where the network predicts residual velocity fields at each iteration. To perform local search in this setting, several modifications are required: the velocity field is first transformed into a deformation field using the scaling and squaring algorithm \cite{ss} with a time step of 5, after which $\boldsymbol{F}_m$ is warped accordingly. The final deformation field is also obtained through the scaling and squaring applied to the accumulated velocity field.

	\renewcommand{\arraystretch}{1.12}
	\begin{table}[!hb]
		\vspace{-3pt}
		\centering
		\setlength{\tabcolsep}{6.5pt}
		\caption[c]{Comparison results on the BTCV dataset without affine pre-registration. MSE serves as the similarity loss for all methods. Arrows ($\uparrow\!\!\!,\downarrow$) indicate the preferable direction of metrics. The standard deviation for each metric is shown in parentheses. \textbf{bold} font indicates the best value, \underline{underline} font indicates the second-best value. Superscripts $^*$ and $^\dagger$ denote different Wilcoxon rank levels against \textit{the best of comparison methods}, denoted in \textcolor{mybrown}{brown}.}
		\vspace{-2pt}
		\scalebox{0.977}{
			\begin{threeparttable}[]
				\begin{tabular}{l|llrr}
					\Xhline{1pt}
					\multirow{4}{*}[0.7ex]{\textbf{Methods}} &  &  &  &  \\[-0.6ex]
					& \multicolumn{4}{c}{\textbf{BTCV} (\textit{w/o Pre-affine}) (MSE)} \\[1.2ex]
					& \multicolumn{1}{c}{Dice\textsubscript{(\%)}$\uparrow$} & \multicolumn{1}{c}{HD95\textsubscript{(mm)}$\downarrow$}  & \multicolumn{1}{c}{ASSD\textsubscript{(mm)}$\downarrow$} & $\%_{\text{fold}}\downarrow$ \\
					\Xhline{0.7pt}    
					\vspace{-7pt}  
					&  &  &  & \\
					\scalebox{0.9}{\quad}Initial & 33.5 \textsubscript{(15.1)} & 41.33 \textsubscript{(15.50)} & 17.69 \textsubscript{(12.52)} & \multicolumn{1}{c}{-}  \\
					\textcolor{orange}{$\circ$} SyN & 61.2 \textsubscript{(11.2)} & 25.57 \textsubscript{(7.70)} & 8.21 \textsubscript{(2.98)} & \underline{1.9$e$-5} \\
					\textcolor{orange}{$\circ$} Demons & 58.0 \textsubscript{(12.4)} & 27.34 \textsubscript{(8.04)} & 8.82 \textsubscript{(3.39)} & 9.0$e$-3 \\
					\textcolor{orange}{$\circ$} B-Spline & 51.6 \textsubscript{(12.0)} & 30.93 \textsubscript{(10.19)} & 10.33 \textsubscript{(3.59)} & 5.8$e$-4 \\
					\textcolor{table_pink}{$\bullet$} VoxelMorph & 48.1 \textsubscript{(9.2)} & 30.76 \textsubscript{(7.35)} & 11.02 \textsubscript{(3.04)} & 7.1$e$-2 \\
					\textcolor{table_pink}{$\bullet$} MIDIR & 46.4 \textsubscript{(8.1)} & 30.85 \textsubscript{(7.19)} & 11.25 \textsubscript{(2.79)} & 1.3$e$-3 \\ 
					\textcolor{table_pink}{$\bullet$} VTN-10 & 52.4 \textsubscript{(10.2)} & 28.95 \textsubscript{(8.33)} & 10.01 \textsubscript{(3.61)} & 2.8$e$-2 \\
					\textcolor{table_pink}{$\bullet$} LapIRN & 56.9 \textsubscript{(9.5)} & 26.41 \textsubscript{(7.19)} & 8.81 \textsubscript{(2.77)} & 5.9$e$-3 \\
					\textcolor{table_pink}{$\bullet$} SDHNet & 63.1 \textsubscript{(11.8)} & 26.31 \textsubscript{(9.42)} & 8.27 \textsubscript{(2.90)} & 5.9$e$-2 \\
					\textcolor{table_pink}{$\bullet$} PCReg & 58.7 \textsubscript{(14.4)} & 28.59 \textsubscript{(7.06)} & 9.51 \textsubscript{(4.84)} & 4.7$e$-2 \\
					\textcolor{table_pink}{$\bullet$} IIRP & 65.5 \textsubscript{(9.3)} & \textbf{\textcolor{mybrown}{24.43}} \textsubscript{(7.06)} & \textcolor{mybrown}{7.39} \textsubscript{(2.52)} & 1.7$e$-2 \\
					\textcolor{table_pink}{$\bullet$} RDP & 64.2 \textsubscript{(10.9)} & 24.64 \textsubscript{(7.86)} & 7.72 \textsubscript{(3.10)} & 5.3$e$-3 \\
					\textcolor{table_yellow}{$\bullet$} TransMorph & 47.0 \textsubscript{(9.9)} & 30.76 \textsubscript{(7.43)} & 11.11 \textsubscript{(3.12)} & 8.4$e$-2 \\
					\textcolor{table_yellow}{$\bullet$} XMorpher & 51.1 \textsubscript{(8.6)} & 28.40 \textsubscript{(7.80)} & 9.98 \textsubscript{(3.05)} & 4.5$e$-2 \\
					\textcolor{table_yellow}{$\bullet$} CGNet & \underline{\textcolor{mybrown}{64.6}} \textsubscript{(9.8)} & 24.51 \textsubscript{(7.74)} & 7.51 \textsubscript{(2.85)} & 2.4$e$-2 \\
					\textcolor{table_qing}{$\bullet$} DualPRNet++ & 59.8 \textsubscript{(9.6)} & 26.73 \textsubscript{(7.59)} & 7.74 \textsubscript{(2.83)} & 2.9$e$-2 \\
					\textcolor{table_qing}{$\bullet$} CorrMLP & 63.7 \textsubscript{(9.5)} & 24.95 \textsubscript{(7.96)} & 7.91 \textsubscript{(2.96)} & 3.2$e$-2 \\
					
					\Xhline{0.5pt}
					&  &  &  &  \\[-2ex]
					\textcolor{table_qing}{$\bullet$} ReCorr-S & 64.0$^*$\textsubscript{(9.1)} & 25.18$^\dagger$\textsubscript{(7.34)} & 7.55$^*$\textsubscript{(2.59)} & 2.0$e$-2 \\			
					\textcolor{table_qing}{$\bullet$} ReCorr & \textbf{66.3}$^\dagger$\textsubscript{(8.9)} & \underline{24.48}$^*$\textsubscript{(7.60)} & \textbf{7.08}$^\dagger$\textsubscript{(2.45)} & 1.9$e$-2 \\
					\textcolor{table_qing}{$\bullet$} ReCorr-S-\itshape{diff} & 63.8$^\dagger$\textsubscript{(9.6)} & 24.83$^\dagger$\textsubscript{(7.32)} & 7.38$^{\ }$\textsubscript{(2.40)} & \textbf{1.1$\boldsymbol{e}$-5}\\
					\textcolor{table_qing}{$\bullet$} ReCorr-\itshape{diff} & 64.5\textsubscript{(10.2)} & 24.82$^\dagger$\textsubscript{(6.97)} & \underline{7.29}$^\dagger$\textsubscript{(2.56)} & 3.0$e$-5\\[0.2ex]
					\Xhline{1pt}
				\end{tabular}
				\vspace{-0.2pt}
				\begin{tablenotes}
					\item[1] {\textcolor{orange}{$\circ$} \itshape{Tranditional Methods}\quad \textcolor{table_pink}{$\bullet$} \itshape{Pure Convolutional Networks}\\ \textcolor{table_yellow}{$\bullet$} \itshape{Region-to-region Methods}\quad\textcolor{table_qing}{$\bullet$} \itshape{Voxel-to-region Methods}}
					\item[2] {\scriptsize $ *: p < 0.05, \quad\dagger: p < 5e$-$5$}
				\end{tablenotes}
			\end{threeparttable}
		}
		\label{tab:btcv}
		\vspace{6pt}
	\end{table}
	\renewcommand{\arraystretch}{1}

	\renewcommand{\arraystretch}{1.12}
	\begin{table*}[!ht]
		\centering
		\setlength{\tabcolsep}{4.2pt}
		\caption[]{Quantitative results on the OASIS dataset, either with affine pre-registration (small deformation) or without affine pre-registration (large deformation). MSE serves as the training similarity loss for all methods. Arrows ($\uparrow\!\!\!,\downarrow$) indicate the preferable direction of metrics. The standard deviation for each metric is shown in parentheses. Average Time and GPU are tested during the inference stage on the non-affine OASIS dataset with the resolution of 192 $\times$ 192 $\times$ 192. \textbf{bold} font indicates the best value,  \underline{underline} font indicates the second-best value. 
		The \textit{best-performing baseline} is highlighted in \textcolor{mybrown}{brown}.  
		Superscripts $^*$ and $^\dagger$ denote different Wilcoxon signed-rank significance levels in comparison with this baseline.
		}
		\vspace{0pt}
		\scalebox{1}{
			\begin{threeparttable}[]
				\begin{tabular}{l|cccccccc|c@{\hskip 6.5pt}c@{\hskip 6.5pt}r@{\hskip 2pt}r}
					\Xhline{1pt}
					\multirow{4}{*}[-1ex]{\textbf{Methods}} &  &  &  &  &  &  &  &  & \multirow{4}{*}[-1ex]{\textbf{Time}} & \multirow{4}{*}[-1ex]{\textbf{GPU}} &  \multirow{4}{*}[-1ex]{\textbf{Params}} & \multirow{4}{*}[-1ex]{\textbf{FLOPs}} \\
					& \multicolumn{8}{c|}{\textbf{OASIS} (MSE)} \\         
					& \multicolumn{4}{c}{\textit{Pre-affine}} & \multicolumn{4}{c|}{\textit{w/o Pre-affine}} \\
					\cmidrule(lr){2-5} \cmidrule(lr){6-9}
					& Dice\textsubscript{(\%)}$\uparrow$ & HD95\textsubscript{(mm)}$\downarrow$ & ASSD \textsubscript{(mm)}$\downarrow$ & $\%_{\text{fold}}\downarrow$ & \multicolumn{1}{c}{Dice\textsubscript{(\%)}$\uparrow$} & \multicolumn{1}{c}{HD95\textsubscript{(mm)}$\downarrow$}  & \multicolumn{1}{c}{ASSD\textsubscript{(mm)}$\downarrow$} & $\%_{\text{fold}}\downarrow$ & (s) & (MB) & & \multicolumn{1}{c}{(GMac)}\\
					\Xhline{0.7pt}    
					\vspace{-7.2pt}   
					&  &  &  &  &  &  &  &  &  &  &   \\
					\scalebox{0.9}{\quad}Initial & 53.8 \textsubscript{(5.7)} & 4.08 \textsubscript{(0.67)} & 1.76 \textsubscript{(0.31)} & - & 13.4 \textsubscript{(10.3)} & 17.00 \textsubscript{(7.81)} & 9.85 \textsubscript{(5.96)} & - & - & - & \multicolumn{1}{c}{-} & \multicolumn{1}{c}{-}\\
					\textcolor{orange}{$\circ$} SyN & 77.5 \textsubscript{(3.1)} & 2.33 \textsubscript{(0.49)} & 0.78 \textsubscript{(0.13)} & \scalebox{0.95}{\textbf{$\boldsymbol{<}$1$\boldsymbol e$-5}} & 69.1 \textsubscript{(3.8)} & 2.67 \textsubscript{(0.51)} & 1.00 \textsubscript{(0.15)} & \scalebox{0.95}{\textbf{$\boldsymbol{<}$1$\boldsymbol e$-5}} & 178 & - & \multicolumn{1}{c}{-} & \multicolumn{1}{c}{-}\\
					\textcolor{orange}{$\circ$} Demons & 77.9 \textsubscript{(2.6)} & 2.37 \textsubscript{(0.45)} & 0.75 \textsubscript{(0.11)} & 3.8$e$-3 & 53.7 \textsubscript{(22.5)} & 5.84 \textsubscript{(5.41)} & 2.62 \textsubscript{(3.25)} & 8.4$e$-4 & 114 & - & \multicolumn{1}{c}{-} & \multicolumn{1}{c}{-}\\
					\textcolor{orange}{$\circ$} B-Spline & 65.0 \textsubscript{(5.4)} & 3.30 \textsubscript{(0.74)} & 1.24 \textsubscript{(0.26)} & \scalebox{0.95}{\textbf{$\boldsymbol{<}$1$\boldsymbol e$-5}} & 60.7 \textsubscript{(6.2)} & 3.42 \textsubscript{(0.83)} & 1.35 \textsubscript{(0.31)}& \scalebox{0.95}{\textbf{$\boldsymbol{<}$1$\boldsymbol e$-5}} & 3.75 & - & \multicolumn{1}{c}{-} & \multicolumn{1}{c}{-}\\
					\textcolor{table_pink}{$\bullet$} VoxelMorph & 78.3 \textsubscript{(2.3)} & 2.08 \textsubscript{(0.31)} & 0.74 \textsubscript{(0.09)} & 2.6$e$-3 & 59.4 \textsubscript{(5.8)} & 3.97 \textsubscript{(1.06)} & 1.45 \textsubscript{(0.30)} & 1.2$e$-3  & 0.13 & 6318 & 0.33 M & 408.5\\
					\textcolor{table_pink}{$\bullet$} MIDIR & 73.6 \textsubscript{(2.5)} & 2.38 \textsubscript{(0.31)} & 0.89 \textsubscript{(0.10)} & \scalebox{0.95}{\textbf{$\boldsymbol{<}$1$\boldsymbol e$-5}} & 55.4 \textsubscript{(5.9)} & 4.22 \textsubscript{(1.13)} & 1.60 \textsubscript{(0.37)} & \scalebox{0.95}{\textbf{$\boldsymbol{<}$1$\boldsymbol e$-5}} & 0.66 & 3712 & 0.21 M & 31.1\\
					\textcolor{table_pink}{$\bullet$} VTN-10 & 77.7 \textsubscript{(1.8)} & 2.13 \textsubscript{(0.31)} & 0.76 \textsubscript{(0.08)} & \underline{1.0$e$-4} & 65.2 \textsubscript{(4.2)} & 3.02 \textsubscript{(0.52)} & 1.14  \textsubscript{(0.19)} & 4.3$e$-4 & 0.83 & 5940 & 2.89 M & 76.2	\\
					\textcolor{table_pink}{$\bullet$} LapIRN & 77.9 \textsubscript{(2.1)} & 2.12 \textsubscript{(0.31)} & 0.75 \textsubscript{(0.09)} & \scalebox{0.95}{\textbf{$\boldsymbol{<}$1$\boldsymbol e$-5}} & 70.1 \textsubscript{(3.3)}& 2.84 \textsubscript{(0.72)} & 1.01 \textsubscript{(0.18)} & \underline{4.4$e$-5} & 0.01 & 5340 & 0.31 M & 288.9\\
					\textcolor{table_pink}{$\bullet$} SDHNet & 79.9 \textsubscript{(2.4)} & 1.87 \textsubscript{(0.28)} & 0.68 \textsubscript{(0.08)} & 3.2$e$-4 & 71.7 \textsubscript{(2.3)} & \textcolor{mybrown}{2.35} \textsubscript{(0.36)} & \textbf{\textcolor{mybrown}{0.77}} \textsubscript{(0.68)} & 1.9$e$-2 & 0.52 & 3510 & 18.29 M & 185.5\\
					\textcolor{table_pink}{$\bullet$} PCReg & 81.3 \textsubscript{(2.2)} & 1.81 \textsubscript{(0.28)} & 0.65 \textsubscript{(0.09)} & 2.2$e$-3 & 74.5 \textsubscript{(2.0)} & 2.95 \textsubscript{(0.48)} & 1.09 \textsubscript{(0.13)} & 5.8$e$-3 & 0.48 & 18530 & 20.09 M & 2264.2\\
					\textcolor{table_pink}{$\bullet$} IIRP & 81.0 \textsubscript{(2.2)} & 1.85 \textsubscript{(0.28)} & 0.66 \textsubscript{(0.10)} & 5.4$e$-4 & 73.9 \textsubscript{(2.3)} & 3.23 \textsubscript{(0.53)} & 1.18 \textsubscript{(0.15)} & 2.3$e$-4 & 3.42 & 4694 & 0.42 M & 2805.6\\
					\textcolor{table_pink}{$\bullet$} RDP & \textcolor{mybrown}{\textbf{81.6}} \textsubscript{(2.1)} & \underline{1.78} \textsubscript{(0.28)} & \underline{0.65} \textsubscript{(0.09)} & \scalebox{0.95}{\textbf{$\boldsymbol{<}$1$\boldsymbol e$-5}} & \textbf{\textcolor{mybrown}{74.8}} \textsubscript{(1.8)} & 2.91 \textsubscript{(0.46)} & 1.08 \textsubscript{(0.12)} & 4.8$e$-5 & 4.54 & 4612 & 8.92 M & 4161.8\\
					\textcolor{table_yellow}{$\bullet$} TransMorph & 80.3 \textsubscript{(2.2)} & 1.88 \textsubscript{(0.27)} & 0.67 \textsubscript{(0.08)} & 2.7$e$-3 & 58.7 \textsubscript{(13.8)} & 5.46 \textsubscript{(4.46)} & 1.56 \textsubscript{(1.59)} & 2.5$e$-3 & 0.20 & 7648 & 46.77 M & 734.2\\
					\textcolor{table_yellow}{$\bullet$} XMorpher & 79.3 \textsubscript{(2.0)} & 2.01 \textsubscript{(0.25)} & 0.73 \textsubscript{(0.07)} & 2.5$e$-3 & 62.1 \textsubscript{(5.1)} & 3.37 \textsubscript{(0.76)} & 1.28 \textsubscript{(0.24)} & 7.8$e$-4 & 1.53 & 6456 & 20.51 M & 1491.2\\
					
					\textcolor{table_yellow}{$\bullet$} CGNet & 81.3 \textsubscript{(2.1)} & 1.79 \textsubscript{(0.28)} & 0.67 \textsubscript{(0.09)} & 1.9$e$-3 & \underline{74.7} \textsubscript{(2.0)} & 2.89 \textsubscript{(0.47)} & 0.83 \textsubscript{(0.47)} & 5.6$e$-3 & 0.57 & 8930 & 25.53 M & 962.2\\
					
					\textcolor{table_qing}{$\bullet$} DualPRNet++ & 79.5 \textsubscript{(2.1)} & 1.89 \textsubscript{(0.29)} & 0.70 \textsubscript{(0.09)} & 2.4$e$-3 & 70.8 \textsubscript{(2.2)} & 3.02 \textsubscript{(0.53)} & 0.97 \textsubscript{(0.61)} & 5.3$e$-3 & 0.48 & 13324 & 0.61 M & 839.8\\
					\textcolor{table_qing}{$\bullet$} CorrMLP & \underline{81.5} \textsubscript{(2.2)} & \textbf{\textcolor{mybrown}{1.77}} \textsubscript{(0.28)} & \textbf{\textcolor{mybrown}{0.64}} \textsubscript{(0.09)} & 2.1$e$-3 & 74.6 \textsubscript{(2.1)} & 2.94 \textsubscript{(0.36)} & 1.09 \textsubscript{(0.68)} & 5.3$e$-3 & 1.03 & 13324 & 4.19 M & 1197.6\\
					
					\Xhline{0.5pt}
					&  &  &  &  &  &  &  &  &  &  & \\[-2.1ex]
					\textcolor{table_qing}{$\bullet$} ReCorr-S & 81.2$^\dagger$\textsubscript{(2.1)} & 1.81$^\dagger$\textsubscript{(0.28)} & \textbf{0.64}$^{\ }$\textsubscript{(0.08)} & 1.6$e$-3  & 74.0$^*$\textsubscript{(2.0)} & \underline{2.30}$^*$\textsubscript{(0.39)} & 0.83$^\dagger$\textsubscript{(0.09)} & 7.8$e$-4 & 0.11 & 5626 & 0.72 M & 243.2\\
					\textcolor{table_qing}{$\bullet$} ReCorr & 81.4$^{\dagger}$\textsubscript{(2.0)} & 1.79$^*$\textsubscript{(0.27)} & \textbf{0.64}$^{\ }$\textsubscript{(0.08)} & 1.6$e$-3  & \underline{74.7}$^{\ }$\textsubscript{(1.9)} & \textbf{2.24}$^\dagger$\textsubscript{(0.37)} & \underline{0.81}$^{\dagger}$\textsubscript{(0.08)} & 7.6$e$-4 & 0.18 & 5626 & 0.72 M & 396.4\\
					\textcolor{table_qing}{$\bullet$} ReCorr-S-\itshape{diff} & 80.6$^\dagger$\textsubscript{(2.0)} & 1.86$^\dagger$\textsubscript{(0.28)} & 0.66$^*$\textsubscript{(0.08)} & \scalebox{0.95}{\textbf{$\boldsymbol{<}$1$\boldsymbol e$-5}} & 71.8$^\dagger$\textsubscript{(2.6)} & 2.40$^{\dagger}$\textsubscript{(0.42)} & 0.88$^\dagger$\textsubscript{(0.11)} & \scalebox{0.95}{\textbf{$\boldsymbol{<}$1$\boldsymbol e$-5}} & 0.24 & 5894 & 0.72 M & 243.4\\
					\textcolor{table_qing}{$\bullet$} ReCorr-\itshape{diff} & 80.9$^\dagger$\textsubscript{(2.0)} & 1.83$^\dagger$\textsubscript{(0.27)} & \underline{0.65}$^\dagger$\textsubscript{(0.08)} &  \scalebox{0.95}{\textbf{$\boldsymbol{<}$1$\boldsymbol e$-5}} & 72.5$^\dagger$\textsubscript{(2.4)} & 2.37$^{* }$\textsubscript{(0.42)} & 0.86$^\dagger$\textsubscript{(0.10)} & \scalebox{0.95}{\textbf{$\boldsymbol{<}$1$\boldsymbol e$-5}} & 0.29 & 6166 & 0.72 M & 396.5\\
					[0.4ex]
					\Xhline{1pt}
				\end{tabular}
				\vspace{-1.2pt}
				\begin{tablenotes}
					\item[1] {\textcolor{orange}{$\circ$} \itshape{Tranditional Methods}\quad \quad\textcolor{table_pink}{$\bullet$} \itshape{Pure Convolutional Methods}\quad\textcolor{table_yellow}{$\bullet$} \itshape{Region-to-region Matching Methods}\quad\textcolor{table_qing}{$\bullet$} \itshape{Voxel-to-region Matching Methods}}
					\item[2] {\scriptsize $ *: p < 0.05, \quad\dagger: p < 5e$-$5$}
				\end{tablenotes}
			\end{threeparttable}
		}
		\label{tab:oasis}
		\vspace{-4pt}
	\end{table*}
	\renewcommand{\arraystretch}{1}

	\vspace{10pt}
	\section{Experiments}
	\vspace{8pt}
	\label{sec:experiment}

	\subsection{Datasets}
	\vspace{0pt}
	The evaluation is performed on 3 datasets: two brain MRI datasets OASIS \cite{oasis} and IXI\footnote{\href{https://brain-development.org/ixi-dataset/}{https://brain-development.org/ixi-dataset/}}, and an abdominal CT dataset BTCV \cite{btcv}. We conduct atlas-based experiments on the OASIS and the IXI datasets, and subject-to-subject experiments on the BTCV dataset. 
	To comprehensively evaluate the effectiveness of ReCorr for both \textit{regular} and \textit{extremely large} deformations, dual-setting experiments are conducted on the OASIS dataset: one with affine pre-registration and the other without. Besides, the IXI dataset preprocessed by TransMorph \cite{transmorph} (including affine pre-registration) is used to test \textit{regular} deformations, while the BTCV dataset without affine pre-registration is used to test \textit{extremely substantial} deformations.
	
	The OASIS dataset includes 414 scans with segmentation annotations on 35 brain regions. One scan is randomly selected as the atlas, and the rest are split into 330 for training, 28 for validation, 55 for testing. For the two experiment settings (with and without affine pre-registration), different preprocessing steps are performed after the skull stripping. Specifically, the OASIS (affine) scans are resampled to a size of $256\times256\times256$ with 1mm isotropic voxels, then affine pre-registerd and cropped to $160\times 224\times 192$. In contrast, the OASIS (non-affine) scans are directly resampled to $192\times 192\times 192$ with 1.33 mm isotropic voxels.

	The IXI dataset includes 576 T1-weighted brain MRI images, and has been preprocessed by TransMorph, with steps include skull stripping, resampling, and affine transformation. All preprocessed images are cropped to a size of $160\times 192\times 224$ and 36 segmentation labels are used for evaluation. The scans are split into 403 training , 58 validation, and 115 test images. Scans from the IXI dataset serve as the moving images and are registered to an atlas brain MRI from \cite{cyclemorph}.
	
	The BTCV dataset includes 50 abdominal multi-organ CT scans from patients with metastatic liver cancer or postoperative abdominal hernia. All scans are resampled to a voxel spacing of $2\times 2\times 2.5$ mm, with intensity values clipped to [-900,\hspace{0.25mm}1000] Hounsfield units and normalized to [0,\hspace{0.25mm}1]. The scans are manually cropped to ensure consistent anatomical coverage and zero-padded to a size of $192\times 160\times 192$. The dataset is split into 35 training, 5 validation, and 10 test scans. For training, each scan is randomly paired with 10 others, resulting in $35 \times  10$ training pairs. Validation and test scans are inter-paired, resulting in $5\times 4$ and $10\times 9$ pairs, respectively. Five organ labels are used for evaluation: the spleen, left kidney, right kidney, liver, and stomach.

	\begin{figure*}[!ht]
		\vspace{-0pt}
		\centerline{\includegraphics[width=1.0\textwidth]{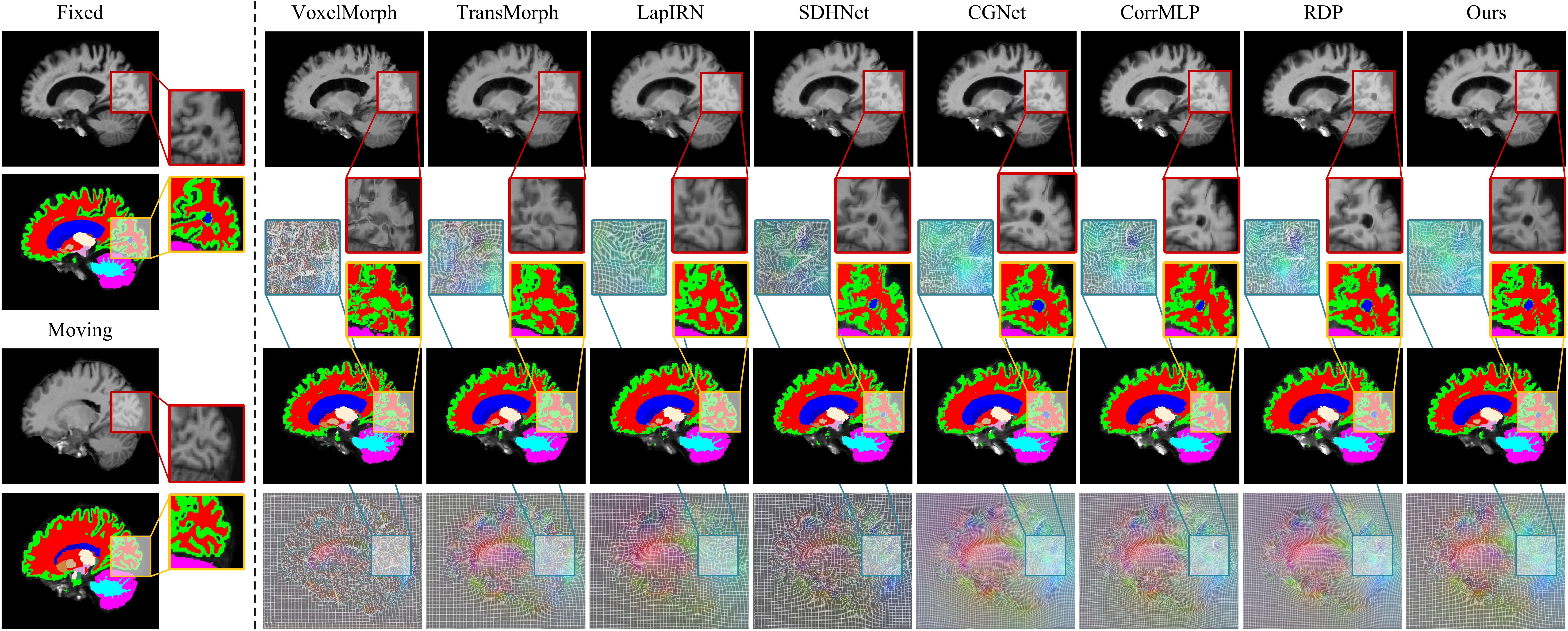}}
		\caption{A visualization example from the OASIS dataset \textit{with affine pre-registration}. Left: the fixed image, the moving image, and their anatomical segmentation, along with zoomed-in views of specific areas at the same location. Right: the 1st, 3rd, and 4th rows show each method's warped moving images, the anatomical segmentation of these images, and the deformation field (shown by RGB images with overlaid grids), respectively. The 2nd row displays zoomed-in views of the same areas in the 1st, 3rd, and 4th rows.}
		\label{fig:oasis}
		\vspace{-1pt}
	\end{figure*}
	
	\renewcommand{\arraystretch}{1.13}
	\begin{table}[!hb]
		\vspace{-11pt}
		\centering
		\setlength{\tabcolsep}{5.6pt}
		\caption[c]{Comparison results on the IXI dataset which has undergone affine pre-registration. NCC serves as the training similarity loss for all methods. Arrows ($\uparrow\!\!\!,\downarrow$) indicate the preferable direction of metrics. The standard deviation for each metric is shown in parentheses. \textbf{bold} font indicates the best value, \underline{underline} font indicates the second-best value. Superscripts $^*$ and $^\dagger$ denote different Wilcoxon rank levels against \textit{the best of comparison methods}, denoted in \textcolor{mybrown}{brown}.}
		\vspace{-1pt}
		\scalebox{1}{
			\begin{threeparttable}[]
				\begin{tabular}{l|cccl}
					\Xhline{1pt}
					\multirow{4}{*}[0.9ex]{\textbf{Methods}} &  &  &  &  \\[-0.8ex]
					& \multicolumn{4}{c}{\textbf{IXI} (\textit{Pre-affine}) (NCC)} \\[1.2ex]
					& Dice\textsubscript{(\%)}$\uparrow$ & HD95 \textsubscript{(mm)} $\downarrow$ & ASSD \textsubscript{(mm)} $\downarrow$ & $\%_{\text{fold}}\downarrow$ \\
					\Xhline{0.7pt}    
					\vspace{-7pt}  
					&  &  &  & \\
					\scalebox{0.9}{\quad}Initial & 35.5 \textsubscript{(3.4)} & 8.00 \textsubscript{(0.78)} & 3.11 \textsubscript{(0.35)} & \multicolumn{1}{c}{-} \\
					\textcolor{orange}{$\circ$} SyN & 65.7 \textsubscript{(2.9)} & 4.97 \textsubscript{(0.50)} & 1.46 \textsubscript{(0.16)} & \multicolumn{1}{c}{\scalebox{0.95}{\textbf{$\boldsymbol{<}$1$\boldsymbol e$-5}}}  \\
					\textcolor{orange}{$\circ$} Demons & 56.8 \textsubscript{(5.3)} & 5.81 \textsubscript{(0.68)} & 1.80 \textsubscript{(0.26)} & 2.8$e$-3  \\
					\textcolor{orange}{$\circ$} B-Spline & 64.0 \textsubscript{(3.6)} & 5.11 \textsubscript{(0.66)} & 1.52 \textsubscript{(0.20)} & \underline{1.1$e$-4}  \\
					\textcolor{table_pink}{$\bullet$} VoxelMorph & 70.6 \textsubscript{(2.4)} & 4.53 \textsubscript{(0.51)} & 1.25 \textsubscript{(0.13)}  & 6.0$e$-3\\
					\textcolor{table_pink}{$\bullet$} MIDIR & 68.1 \textsubscript{(2.4)} & 4.71 \textsubscript{(0.52)} & 1.36 \textsubscript{(0.15)}  & \multicolumn{1}{c}{\scalebox{0.95}{\textbf{$\boldsymbol{<}$1$\boldsymbol e$-5}}} \\
					\textcolor{table_pink}{$\bullet$} VTN-10 & 70.5 \textsubscript{(2.2)} & 4.70 \textsubscript{(0.50)} & 1.27 \textsubscript{(0.13)} & 7.0$e$-3  \\
					\textcolor{table_pink}{$\bullet$} LapIRN & 71.2 \textsubscript{(2.3)} & \textcolor{mybrown}{4.44} \textsubscript{(0.49)} & 1.23 \textsubscript{(0.13)} & 2.7$e$-4 \\
					\textcolor{table_pink}{$\bullet$} SDHNet & 70.4 \textsubscript{(3.4)} & 4.59 \textsubscript{(0.49)} & 1.26 \textsubscript{(0.12)} & 1.3$e$-2  \\
					\textcolor{table_pink}{$\bullet$} PCReg & 70.4 \textsubscript{(2.1)} & 4.52 \textsubscript{(0.49)} & 1.57 \textsubscript{(0.18)} & 4.0$e$-3  \\
					\textcolor{table_pink}{$\bullet$} IIRP & 71.6 \textsubscript{(1.9)} & \textcolor{mybrown}{4.44} \textsubscript{(0.48)} & 1.54 \textsubscript{(0.17)} & 1.9$e$-3  \\
					\textcolor{table_pink}{$\bullet$} RDP & \textcolor{mybrown}{71.4} \textsubscript{(2.1)} & 4.47 \textsubscript{(0.49)} & 1.53 \textsubscript{(0.18)} & 2.7$e$-4  \\
					\textcolor{table_yellow}{$\bullet$} TransMorph & 70.5 \textsubscript{(2.4)} & 4.52 \textsubscript{(0.51)} & 1.24 \textsubscript{(0.13)} & 6.3$e$-3 \\
					\textcolor{table_yellow}{$\bullet$} XMorpher & 70.7 \textsubscript{(2.3)} & 4.49 \textsubscript{(0.48)} & 1.24 \textsubscript{(0.13)} & 5.1$e$-3 \\
					\textcolor{table_yellow}{$\bullet$} CGNet & 70.5 \textsubscript{(2.2)} & 4.57 \textsubscript{(0.49)} & \textcolor{mybrown}{1.21} \textsubscript{(0.52)} & 4.3$e$-3  \\
					\textcolor{table_qing}{$\bullet$} DualPRNet++ & 70.1 \textsubscript{(3.2)} & 4.63 \textsubscript{(0.49)} & 1.26 \textsubscript{(0.12)} & 2.1$e$-2  \\
					\textcolor{table_qing}{$\bullet$} CorrMLP & 70.4 \textsubscript{(3.4)} & 4.59 \textsubscript{(0.50)} & 1.25 \textsubscript{(0.15)} & 1.3$e$-2  \\
					
					\Xhline{0.5pt}
					&  &  &  &  \\[-2ex]
					\textcolor{table_qing}{$\bullet$} ReCorr-S & \underline{72.5}$^\dagger$\textsubscript{(1.8)} & \underline{4.37}$^\dagger$\textsubscript{(0.48)} & \underline{1.19}$^\dagger$\textsubscript{(0.12)} & 3.4$e$-3 \\
					\textcolor{table_qing}{$\bullet$} ReCorr & \textbf{72.6}$^\dagger$\textsubscript{(1.8)} & \underline{4.37}$^\dagger$\textsubscript{(0.47)} & \textbf{1.18}$^\dagger$\textsubscript{(0.12)} & 2.9$e$-3 \\
					\textcolor{table_qing}{$\bullet$} ReCorr-S-\itshape{diff} & 72.1$^\dagger$\textsubscript{(2.0)} & 4.38$^\dagger$\textsubscript{(0.48)} & 1.20$^\dagger$\textsubscript{(0.11)} & \scalebox{0.95}{\textbf{$\boldsymbol{<}$1$\boldsymbol e$-5}}\\
					\textcolor{table_qing}{$\bullet$} ReCorr-\itshape{diff} & 72.2$^\dagger$\textsubscript{(1.9)} & \textbf{4.36}$^\dagger$\textsubscript{(0.48)} & \underline{1.19}$^\dagger$\textsubscript{(0.12)} & 1.2$e$-5\\[0.2ex]
					\Xhline{1pt}
				\end{tabular}
				\begin{tablenotes}
					\item[1] {\textcolor{orange}{$\circ$} \itshape{Tranditional Methods}\quad \textcolor{table_pink}{$\bullet$} \itshape{Pure Convolutional Networks}\\ \textcolor{table_yellow}{$\bullet$} \itshape{Region-to-region Methods}\quad\textcolor{table_qing}{$\bullet$} \itshape{Voxel-to-region Methods}}
					\item[2] {\scriptsize $ *: p < 0.05, \quad\dagger: p < 5e$-$5$}
				\end{tablenotes}
			\end{threeparttable}
		}
		\label{tab:ixi}
		\vspace{-1pt}
	\end{table}
	\renewcommand{\arraystretch}{1}
	
	\vspace{-4pt}
	\subsection{Evaluation Metrics}
	\vspace{-1pt}
	\tred{We evaluate registration accuracy using} Dice coefficient, 95\% Hausdorff Distance (HD95), and Average Symmetric Surface Distance (ASSD). Dice quantifies the overlap between anatomical segmentations \tred{(higher is better,} maximum of $1$). For multiple labels, we calculate an average Dice score. HD95 calculates the 95\% largest distance between the closest points of two anatomical outlines \tred{(lower is better)}, offering more robustness against outliers than standard Hausdorff Distance. ASSD measures the average point-to-point surface distance, which provides a more comprehensive perspective and serves as a balanced quality metric. HD95 and ASSD are reported in millimeters (mm), computed using the voxel spacing information provided with each image.
	
	Additionally, \tred{we assess deformation smoothness} using the determinant of the Jacobian matrix, $|J_{\phi}(\boldsymbol{p})|=|\nabla\phi(\boldsymbol{p})|\in\mathbb R^{3\times 3}$, which reflects the local characteristics of $\phi$ around voxel $\boldsymbol{p}$ where negative values suggest image folding. The percentage of non-positive values in $|J_{\phi}(\boldsymbol{p})|$, denoted as $\%_{\text{fold}}$, is used to quantify folding. Besides, we evaluate inference time, inference memory usage, parameter number, and FLOPs. For traditional methods, \tred{runtime is measured on CPU, while deep learning modles are tested} on an NVIDIA L40 GPU.
	
	To assess statistical significance, we conduct Wilcoxon signed-rank tests between our method and the best-performing comparison method under each metric, using per-patient scores.

	\vspace{-5pt}
	\subsection{Baseline Methods}
	\vspace{-1pt}
	
	\subsubsection{Traditional Methods}
	We select three widely-used algorithms for traditional methods: the Symmetric Normalization (SyN) \cite{SyN}, Demons \cite{demons}, and B-Spline \cite{bspline}. Standard SyN is implemented using the ANTsPy Python package, with iterations set to (160, 80, 40). Demons registration is executed using the SimpleITK \cite{sitk} Python package in a multi-scale fashion, with the iteration number set to 10. The B-Spline algorithm is applied using elastix \cite{elastix}, with iterations maximized at 500. For both SyN and B-Spline, the choice of metric is dataset-specific: MSE for the OASIS and BTCV datasets, and NCC for the IXI dataset, whose atlas and subjects exhibit wide variance in contrast and intensity distribution.

	\begin{figure*}[!ht]
		\vspace{-4pt}
		\centerline{\includegraphics[width=0.999\textwidth]{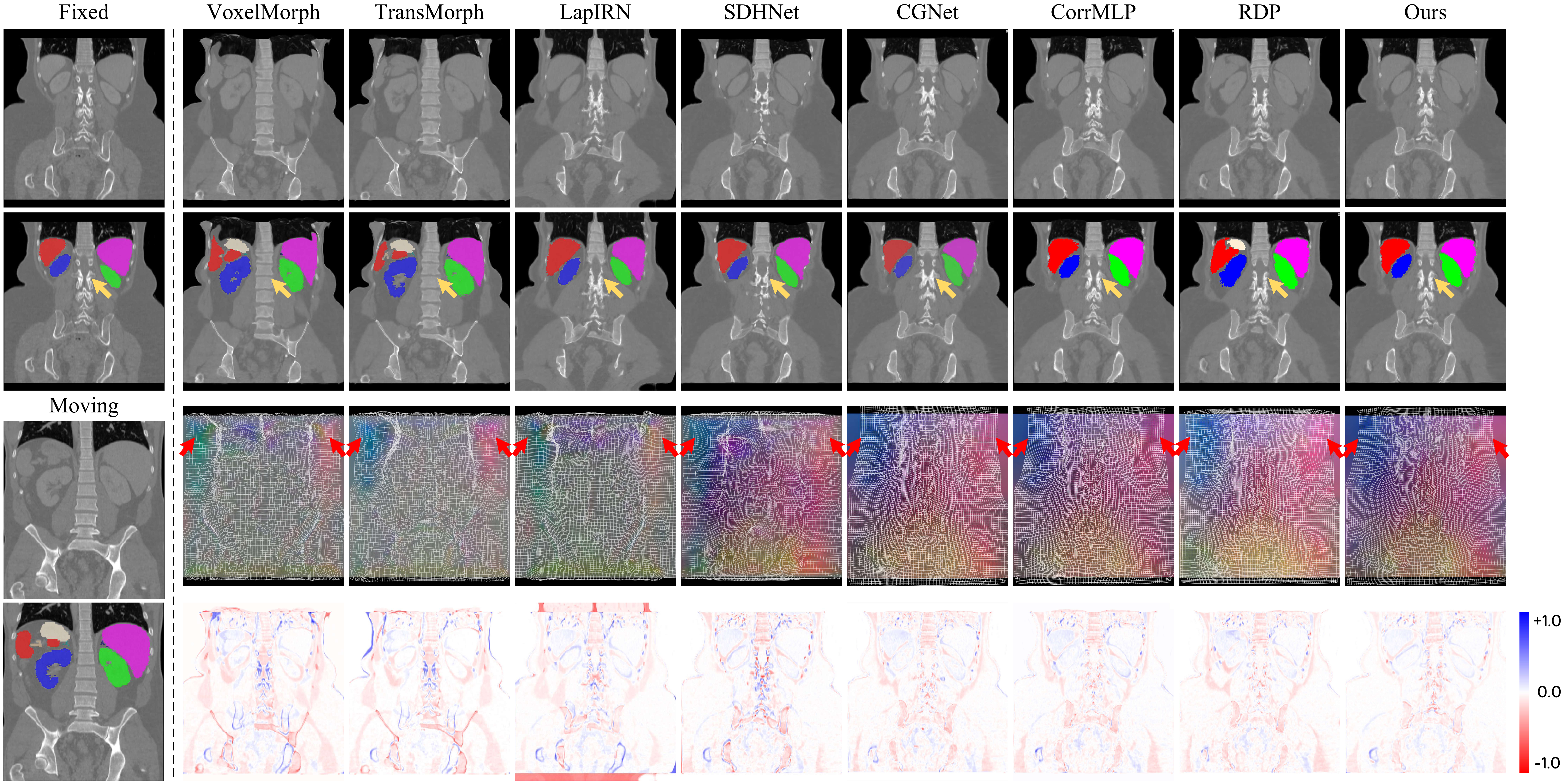}}
		\caption{A visualization example in the BTCV dataset without affine pre-registration. The first column shows the fixed and moving images with their segmentations; the remaining columns show results from different methods. The rows display, from top to bottom: warped images, warped segmentations, deformation fields (RGB grids), and difference maps.
		Yellow arrows indicate a location to distinguish performance. Red arrows indicate the locations where the deformation field generated by our method can demonstrate large linear stretching. }
		\label{fig:btcv}
		\vspace{-0pt}
	\end{figure*}
	
	\vspace{2pt}
	\subsubsection{Learning-based Methods}
	We reimplement thirteen learning-based approaches across three categories: pure convolutional methods, region-to-region explicit matching methods, and voxel-to-region explicit matching methods. 
	
	For pure convolutional methods, we include VoxelMorph \cite{voxelmorph} and MIDIR \cite{midir} as direct registration baselines; VTN \cite{vtn} with 10 cascaded subnetworks as a representative recursive method; LapIRN \cite{lapirn} and PCReg \cite{pcreg} as representative pyramid-based methods; SDHNet \cite{sdhnet} (with 6 iterations), RDP \cite{rdp} and IIRP \cite{iirp} as examples of pyramid-recurrent networks. 
	For region-to-region matching methods, we compare TransMorph \cite{transmorph}, which integrates both Transformer and convolution modules; XMorpher \cite{xmorpher}, a fully Transformer-based model; and CGNet \cite{cgnet}, which adopts a pyramid structure with cross-window correlation.
	For voxel-to-region matching methods, we include DualPRNet++ \cite{dualprnetplus}, and CorrMLP \cite{corrmlp}, which expands the receptive field using multi-window MLPs.

	For all the comparison methods, MSE is used as the similarity loss on the OASIS and BTCV datasets, with a regularization weight of $0.02$. For the IXI dataset, due to substantial differences in data distribution, we choose NCC with a window size of 9 and a regularization weight of $1$. The learning rate is set to $0.0001$ with a weight decay rate of $0.0001$, and the batch size is 1. Each method is trained for 200 epochs with validation performed after each epoch. The model achieving the best performance on the validation set is selected  for final testing.

	\vspace{-2pt}
	\subsection{Implementation Details}
	\vspace{-0pt}
	
	ReCorr is implemented in PyTorch and training using the AdamW optimizer with a learning rate of $0.0007$, a weight decay rate of $0.0004$ and a batch size of 1. Consistent with other learning-based methods, our method is trained for 200 epochs. 
	ReCorr performs \{3,3,2,2\} recurrent iterations across four resolution scales (\scalebox{1.25}{\sfrac{1}{16}, \sfrac{1}{8}, \sfrac{1}{4}, \sfrac{1}{2}}). To accelerate inference in scenarios with relatively small deformations, we also provide ReCorr-S, a faster variant that uses fewer iterations \{1,1,1,1\} at each scale. The diffeomorphic versions ReCorr-\textit{diff} and ReCorr-S-\textit{diff} are implemented under the same configurations.
	
	The similarity loss function and its corresponding regularization weight are selected to match those used in other comparison methods: MSE with $\lambda$ = 0.02 for the OASIS and BTCV datasets, and NCC with $\lambda$ = 1 for the IXI dataset. The iteration decay hyperparameter $\gamma$ is set to $0.7$.

	\vspace{8pt}
	\section{Results}
	\label{sec:results}
	\vspace{4pt}
	
	To demonstrate the efficiency and accuracy of our method across different deformation scales, we conduct comprehensive comparisons with the existing methods in two settings: 1) \textit{Regular} experiments with affine pre-registration for small deformations, and 2) \textit{Extreme} tests on datasets without affine pre-registration for large deformations.

	\vspace{-5pt}
	\subsection{Compare to Start-of-the-art Methods}
	\vspace{0pt}
	
	We evaluate our method across four benchmark datasets: OASIS (pre-affine and non-affine), IXI, and BTCV. The quantitative results are shown in Tables \ref{tab:oasis}, \ref{tab:ixi} and \ref{tab:btcv}, our proposed ReCorr consistently achieves competitive or superior accuracy compared with recent state-of-the-art methods, while offering a strong balance between accuracy and efficiency.
	
	On the non-affine OASIS and BTCV datasets, which involve large anatomical deformations, ReCorr achieves strong results in all metrics. 
	On BTCV, ReCorr achieves the highest Dice score (66.3\%) as well as the best HD95 (24.48~mm) and ASSD (7.08~mm) among all methods, confirming its advantage under large-deformation conditions. While on OASIS (non-affine), ReCorr reaches a Dice score of 74.7\%, ranking second only to RDP (74.8\%), but yields a lower HD95 (2.24~mm \textit{vs.} 2.91~mm) and significantly better efficiency in terms of parameters (0.72M \textit{vs.} 8.92M), FLOPs (396.4G \textit{vs.} 4161.8G), and inference time (0.18~s \textit{vs.} 4.54~s). In contrast to most methods with comparable accuracy, ReCorr achieves these results with only 0.72M parameters and moderate computational cost, demonstrating a strong accuracy-efficiency trade-off under large-deformation scenarios. 
	Fig. \ref{fig:trade-off} visualizes the relationship between registration accuracy (Dice), inference time, FLOPs, and model parameters on the non-affine OASIS dataset. ReCorr achieves one of the highest Dice scores while maintaining moderate inference time, FLOPs, and parameter count. Compared to methods with similar accuracy, such as RDP and CorrMLP, ReCorr requires substantially fewer computational resources. Overall, our method achieves an excellent accuracy-efficiency trade-off.
	
	On the pre-affine datasets OASIS and IXI, where deformations are relatively mild, our method remains highly competitive. ReCorr-S achieves Dice scores of 81.2\% on OASIS and 72.5\% on IXI, performing on par with or surpassing most existing methods. On OASIS, some approaches such as CorrMLP (81.5\%) and RDP (81.6\%) report slightly higher Dice scores, but incur significantly greater computational cost. In contrast, ReCorr-S achieves comparable accuracy with only 0.72M parameters, 243G FLOPs, and a fast 0.11s inference time. On the IXI dataset, ReCorr-S not only surpasses all previous methods in Dice, outperforming RDP (71.4\%) and IIRP (71.6\%), but also maintains the lowest computational burden among top-performing methods. These results highlight the strong accuracy-efficiency trade-off of ReCorr-S under small-deformation conditions, making it a practical and effective choice for time- and resource-sensitive scenarios.

	\renewcommand{\arraystretch}{1.12}
	\begin{table}[!b]
		\vspace{-2pt}
		\centering
		\setlength{\tabcolsep}{6.2pt}
		\caption[c]{Comparison results under the semi-supervised setting on the non-affine OASIS dataset and IXI dataset. Arrows ($\uparrow\!\!\!,\downarrow$) indicate the preferable direction of metrics. The standard deviation for each metric is shown in parentheses. \textbf{Bold} font indicates the best value, \underline{underline} font indicates the second-best value.}
		\vspace{-2pt}
		\scalebox{0.977}{
			\begin{threeparttable}[]
				\begin{tabular}{l|llcc}
					\Xhline{1pt}
					\vspace{-5pt}
					& & & & \\[-1ex]
					\multirow{4}{*}[0.7ex]{\textbf{Methods}} &  &  &  &  \\[-0.6ex]
					& \multicolumn{2}{c}{\textit{non-affine} \textbf{OASIS} (MSE)} & \multicolumn{2}{c}{\textbf{IXI} (NCC)}\\[1.2ex]
					\cmidrule(lr){2-3}\cmidrule(lr){4-5}
					& \multicolumn{1}{c}{Dice\textsubscript{(\%)}$\uparrow$} & \multicolumn{1}{c}{HD95\textsubscript{(mm)}$\downarrow$}  & \multicolumn{1}{c}{Dice\textsubscript{(\%)}$\uparrow$} & \multicolumn{1}{c}{HD95\textsubscript{(mm)}$\downarrow$} \\
					\Xhline{0.7pt}    
					\vspace{-7pt}   
					&  &  &  & \\
					\scalebox{0.9}{\quad}Initial & 13.4 \textsubscript{(10.3)} & 17.00 \textsubscript{(7.81)} & 35.5 \textsubscript{(3.4)} & 8.00 \textsubscript{(0.78)}  \\
					\textcolor{table_pink}{$\bullet$} VoxelMorph & 73.8 \textsubscript{(2.2)}  & 3.58 \textsubscript{(0.57)}  & 82.1 \textsubscript{(1.5)} & 3.71 \textsubscript{(0.44)} \\
					\textcolor{table_pink}{$\bullet$} MIDIR  & 71.9 \textsubscript{(5.2)}  & 3.69 \textsubscript{(1.01)}  & 79.4 \textsubscript{(1.7)} & 3.92 \textsubscript{(0.47)} \\
					\textcolor{table_pink}{$\bullet$} VTN-10 & 85.1 \textsubscript{(1.8)}  & 2.24 \textsubscript{(0.47)}  & 80.2 \textsubscript{(1.6)} & 3.84 \textsubscript{(0.45)} \\
					\textcolor{table_pink}{$\bullet$} SDHNet & 86.2 \textsubscript{(1.3)}  &  1.90 \textsubscript{(0.29)} & 83.6 \textsubscript{(1.6)} & 3.61 \textsubscript{(0.46)} \\
					\textcolor{table_pink}{$\bullet$} PCReg  & 87.2 \textsubscript{(1.2)}  & 2.01 \textsubscript{(0.38)}  & 83.6 \textsubscript{(1.7)} & 3.59 \textsubscript{(0.45)} \\
					\textcolor{table_pink}{$\bullet$} IIRP   & 86.4 \textsubscript{(1.1)}  & 1.98 \textsubscript{(0.32)}  & 83.7 \textsubscript{(1.8)} & \underline{3.55} \textsubscript{(0.42)} \\
					\textcolor{table_pink}{$\bullet$} RDP    & \textbf{89.1} \textsubscript{(1.2)}  & 2.09 \textsubscript{(0.41)} & \underline{84.3} \textsubscript{(1.4)} & 3.60 \textsubscript{(0.42)} \\
					\textcolor{table_yellow}{$\bullet$} TransMorph  & 74.3 \textsubscript{(5.4)}  & 3.74 \textsubscript{(1.27)}  & 82.1 \textsubscript{(1.5)} & 3.71 \textsubscript{(0.41)} \\
					\textcolor{table_yellow}{$\bullet$} XMorph  & 79.6 \textsubscript{(1.5)}  & 2.85 \textsubscript{(0.55)}  & 83.3 \textsubscript{(1.7)} & 3.66 \textsubscript{(0.44)} \\
					\textcolor{table_yellow}{$\bullet$} CGNet   & 88.7 \textsubscript{(1.2)}  & \underline{1.88} \textsubscript{(0.33)}  & 82.9 \textsubscript{(1.5)} & 3.67 \textsubscript{(0.44)} \\
					\textcolor{table_qing}{$\bullet$} DualPRNet++ & 88.3 \textsubscript{(1.3)}  & 1.96 \textsubscript{(0.42)}  & 83.2 \textsubscript{(1.6)} & 3.65 \textsubscript{(0.43)} \\
					\textcolor{table_qing}{$\bullet$} CorrMLP & 88.3 \textsubscript{(1.2)}  & 1.92 \textsubscript{(0.34)}  & 83.7 \textsubscript{(1.8)} & \underline{3.55} \textsubscript{(0.45)} \\[0.2ex]
					\Xhline{0.5pt}
					&  &  &  &  \\[-2.2ex]
					\textcolor{table_qing}{$\bullet$} ReCorr & \underline{88.9} \textsubscript{(1.2)}  & \textbf{1.82} \textsubscript{(0.30)}  & \textbf{85.8} \textsubscript{(1.6)} & \textbf{3.52} \textsubscript{(0.43)} \\[0.2ex]
					\Xhline{1pt}
				\end{tabular}
				\vspace{-0.2pt}
				\begin{tablenotes}
					\item[1] {\textcolor{orange}{$\circ$} \itshape{Tranditional Methods}\quad \textcolor{table_pink}{$\bullet$} \itshape{Pure Convolutional Networks}\\ \textcolor{table_yellow}{$\bullet$} \itshape{Region-to-region Methods}\quad\textcolor{table_qing}{$\bullet$} \itshape{Voxel-to-region Methods}}
				\end{tablenotes}
			\end{threeparttable}
		}
		\label{tab:semi}
		\vspace{1pt}
	\end{table}
	\renewcommand{\arraystretch}{1}
	
	Fig. \ref{fig:oasis}. visualizes the comparison results of ReCorr-S with some representative methods of pure convolutional methods, region-to-region matching methods, and voxel-to-region matching methods. Compared to other methods, our approach produces a warped image with more accurate details, while maintaining smoother and more coherent deformation fields.
	Fig.~\ref{fig:btcv} shows qualitative results of ReCorr on the non-affine BTCV dataset. As seen in the third row, our model effectively captures rigid, linear displacements from the moving to fixed image, despite not explicitly modeling affine transformations. Additionally, the deformation fields generated by our method are visually smoother than those from other methods.

	\renewcommand{\arraystretch}{1.16}
	\begin{table}[]
		\caption{Ablation results on Local Search. Settings used in our final model are underlined. Bold font indicates the best value.}
		\setlength{\tabcolsep}{2.8pt}
		\begin{tabular}{lcclcc}
			\Xhline{1pt}
			&  &  &  \\[-1.4ex]
			\multirow{4}{*}[2.6ex]{\textbf{Experiment}} & \multicolumn{2}{c}{\textit{non-affine} \textbf{OASIS}} &  \multicolumn{2}{c}{\textbf{BTCV}} & \textbf{FLOPs}\\
			\cmidrule(lr){2-3}\cmidrule(lr){4-5}
			
			& Dice\textsubscript{(\%)}$\uparrow$  & HD95 \textsubscript{(mm)}$\downarrow$   & Dice\textsubscript{(\%)}$\uparrow$  & HD95 \textsubscript{(mm)}$\downarrow$ & (GMac)\\ 
			\Xhline{0.7pt}
			\vspace{-8.5pt}
			&  &  &  \\
			feature only & 72.9 (2.1)  & 2.38 (0.43) & 55.4 (10.7) & 28.85 (7.56) & 396.2\\
			r=1(corr) & 73.5 (2.1) & 2.34 (0.42) & 63.6 (9.3) & 25.75 (7.32) & 395.5 \\
			\underline{r=3(corr)} & \textbf{74.7} (1.9) & \textbf{2.24} (0.37) & 66.3 (8.9) & \textbf{24.48} (7.60) & 396.4\\
			r=5(corr)  & 74.7 (1.9) & 2.24 (0.36) & \textbf{66.4} (9.0) & 24.53 (7.69) & 399.5\\[0.1ex]
			\Xhline{1pt}
		\end{tabular}
		\label{tab:local search}
	\end{table}
	\renewcommand{\arraystretch}{1}

	\renewcommand{\arraystretch}{1.16}
	\begin{table}[]
		\setlength{\tabcolsep}{3pt}
		\vspace{2pt}
		\caption{Ablation results on the Recurrent Updater. Bold font indicates the best value.}
		\begin{tabular}{c@{\hskip 2.5pt}c@{\hskip 2pt}|c@{\hskip 2pt}c@{\hskip 2pt}c@{\hskip 2pt}|c@{\hskip 2pt}c@{\hskip 1pt}|c@{\hskip 3pt}l}
			\Xhline{1pt}
			&  &  &  &  &  &  &  &  \\[-1.4ex]
			\multicolumn{2}{c|}{Feature} & \multicolumn{3}{c|}{Components} &
			\multicolumn{2}{c|}{Param. Share} &
			\multicolumn{2}{c}{\textit{non-affine} \textbf{OASIS}}  \\
			\cmidrule(lr){8-9}
			Decouple & Concat & Conv & GRU & LSTM & Scale & All & Dice\textsubscript{(\%)}$\uparrow$  & HD95 \textsubscript{(mm)}$\downarrow$   \\ 
			\Xhline{0.7pt}    
			\vspace{-7.2pt}   
			&  &  &  &  &  &  &  &   \\
			& \scalebox{1.15}{\ding{51}}  &  & \scalebox{1.15}{\ding{51}} &  & \scalebox{1.15}{\ding{51}}  &  & 73.6 \textsubscript{(2.2)} & 3.21 \textsubscript{(0.57)} \\
			\scalebox{1.15}{\ding{51}} & & \scalebox{1.15}{\ding{51}} &  &  & \scalebox{1.15}{\ding{51}} &  &  74.1 \textsubscript{(2.0)}  & 2.26 \textsubscript{(0.38)}    \\
			\scalebox{1.15}{\ding{51}} & &  &  & \scalebox{1.15}{\ding{51}} & \scalebox{1.15}{\ding{51}} &  &  74.5 \textsubscript{(1.9)}  & 2.29 \textsubscript{(0.47)}   \\
			\scalebox{1.15}{\ding{51}} & &  & \scalebox{1.15}{\ding{51}} &  &  & \scalebox{1.15}{\ding{51}} & 74.4 \textsubscript{(2.0)} & 2.25 \textsubscript{(0.37)}   \\
			\scalebox{1.15}{\ding{51}} & &  & \scalebox{1.15}{\ding{51}} &  & \scalebox{1.15}{\ding{51}} &  & \textbf{74.7} \textsubscript{(1.9)}  & \textbf{2.24} \textsubscript{(0.37)}  \\[0.1ex]
			\Xhline{1pt}
		\end{tabular}
		\label{tab:update}
		\vspace{-4pt}
	\end{table}
	\renewcommand{\arraystretch}{1}

	\vspace{-2pt}
	\subsection{Trade-off between ReCorr and ReCorr-S}

	While both ReCorr and ReCorr-S are built upon the same framework, their performance trends across datasets reveal practical insights. On datasets with small deformations (\textit{i.e.}, pre-affine OASIS and IXI), the performance gap is marginal: ReCorr-S achieves nearly the same Dice scores and in some cases even slightly lower HD95, suggesting that a single iteration per scale is sufficient to achieve accurate alignment. Given their identical parameter count, ReCorr-S is preferred in such cases due to its shorter inference time.
	
	In contrast, on datasets with large deformations (\textit{i.e.}, non-affine OASIS and BTCV), ReCorr shows clearer improvements over ReCorr-S. It achieves a 0.7\%–2.3\% higher Dice score, along with noticeably improved HD95 and ASSD, demonstrating the benefit of additional recurrent updates in handling complex anatomical variability. These observations suggest a flexible usage strategy: ReCorr-S is well-suited for efficiency-critical applications and performs effectively under small deformations, while ReCorr is better suited for scenarios requiring maximum accuracy in large-deformation cases.

	\vspace{-7pt}
	\subsection{Semi-supervised Setting}
	\label{sec:semi}
	\vspace{-0pt}
	Optionally, to further improve performance, we adopt a semi-supervised training strategy following \cite{voxelmorph}, where an additional Dice loss is used on labeled data alongside the unsupervised similarity and regularization losses. We evaluate this setup on the non-affine OASIS and IXI datasets, which contain varying degrees of deformation.
	
	As shown in Table~\ref{tab:semi}, ReCorr achieves state-of-the-art performance under this setting. On the OASIS dataset, ReCorr attains a Dice score of 88.9\%, surpassing the previous best result (89.1\% by RDP) with a notably lower HD95 (1.82~mm vs. 2.09~mm). On the IXI dataset, which exhibits smaller deformations, ReCorr reaches the highest Dice score (85.8\%) and the lowest HD95 (3.52~mm) among all methods. These results demonstrate that our model benefits from the additional supervised signal and remains robust across both small and large deformation scenarios.

	\vspace{-8pt}
	\subsection{Ablation Study}
	\label{sec:ablation}
	\vspace{-1pt}

	Our ablation experiments assess the effectiveness of the local search module, recurrent updater, multi-scale architecture, operations at the original resolution, and sequence supervision strategy, as well as the hyperparameters of iteration numbers and $\gamma$. Unless specifically emphasized, we default to conducting ablation experiments for ReCorr on the non-affine OASIS dataset with Dice and HD95.
	
	\vspace{1pt}
	\subsubsection{Local Search} 
	
	We conduct ablation studies to assess the design choices in the local search module, as shown in Table \ref{tab:local search}. The ``feature only" variant skips correlation computation and directly uses concatenated features from the fixed and moving images. Using correlation yields clear improvements (e.g., 74.7\% vs. 72.9\% Dice on non-affine OASIS), indicating its advantage in capturing voxel correspondences. We further evaluate different search scopes ($r$=1, 3, 5) using correlation-based matching. Increasing the scope substantially improves performance from $r$=1 to $r$=3, especially on the BTCV dataset with large deformations. The performance at $r=5$ shows no clear improvement over $r=3$ (e.g., 66.4\% vs. 66.3\% Dice on BTCV), and the computational cost remains nearly the same. This suggests that $r=3$ already provides sufficient search flexibility, and further increasing the radius offers little benefit.

	\renewcommand{\arraystretch}{1.13}
	\begin{table}[!t]
		\vspace{-0pt}
		\centering
		\setlength{\tabcolsep}{7.3pt}
		\caption[c]{Ablation studies for iteration numbers on 4 scales. The bold font indicates the best value.}
		\vspace{-2pt}
		\scalebox{1}{
			\begin{tabular}{cccccc}
				\Xhline{1pt}
				&  &  &  &  &  \\[-1ex]
				\multicolumn{4}{c}{\textbf{Iterations}} & \multicolumn{2}{c}{\textbf{OASIS} (\textit{w/o Pre-affine)}} \\
				\cmidrule(lr){5-6}
				1/16       & 1/8 & 1/4 & 1/2 & Dice \textsubscript{(\%)}$\uparrow$ & HD95 \textsubscript{(mm)}$\downarrow$  \\
				\Xhline{0.7pt}    
				\vspace{-7pt}  
				&  &  &  &  &  \\
				0  & 1  & 0  & 0  & 59.9 (5.4) & 3.65 (0.83) \\
				0  & 1  & 1  & 0  & 70.3 (2.9) & 2.69 (0.52) \\
				1  & 1  & 1  & 0  & 71.6 (2.0) & 2.50 (0.38)\\
				1  & 1  & 1  & 1  & \textbf{74.0} (2.0) & \textbf{2.30} (0.39) \\
				\Xhline{0.7pt}
				\vspace{-7pt}
				&  &  &  \\[-0.5ex]
				2  & 1  & 1  & 1  & 74.3 (1.9) & 2.26 (0.37) \\
				1  & 2  & 1  & 1  & 74.3 (2.0) & 2.27 (0.38)  \\
				1  & 1  & 2  & 1  & 74.2 (1.9) & 2.27 (0.36) \\
				1  & 1  & 1  & 2  & 74.4 (1.9) & 2.26 (0.37) \\
				2  & 2  & 2  & 1  & 74.3 (2.0) & 2.26 (0.38) \\
				3  & 3  & 1  & 1  & 74.5 (2.0) & 2.25 (0.37) \\
				2  & 2  & 2  & 2  & 74.6 (2.0) & 2.26 (0.41) \\
				2  & 3  & 2  & 2  & 74.6 (1.9) & 2.25 (0.38) \\
				3  & 3  & 2  & 2  & \textbf{74.7} (1.9) & \textbf{2.24} (0.37) \\
				\Xhline{1pt}
			\end{tabular}
		}
		\label{tab:ab:iteration}
		\vspace{-2pt}
	\end{table}
	\renewcommand{\arraystretch}{1}

	\renewcommand{\arraystretch}{1.1}
	\begin{table}[!t]
		\centering
		\vspace{1pt}
		\setlength{\tabcolsep}{4.2pt}
		\caption[c]{Ablation studies for operations at original resolution. Settings used in our final model are underlined. The bold font indicates the best value.}
		\scalebox{1}{
			\begin{tabularx}{0.98\linewidth}{c@{\hskip 5pt}c@{\hskip 3pt}c|ccccc}
				
				\Xhline{1pt}
				\vspace{-2pt}
				&  &  &  &  &  &\\[-0.6ex]
				\multicolumn{3}{c|}{\textbf{Ori-Res. Setting}} & \multicolumn{2}{c}{\textbf{OASIS} \textit{(w/o Pre-affine)}} & GPU & Time & FLOPs\\
				&  &  &  &  &  &\\[-2.1ex]
				Plain & \underline{Refine} & Iter.(1) & Dice \textsubscript{(\%)}$\uparrow$ & HD95 \textsubscript{(mm)}$\downarrow$ & (MB) & (s) & (GMac) \\
				\Xhline{0.7pt}    
				&  &  &  &  &  & \\[-2ex]
				\scalebox{1.15}{\ding{51}} & & & 74.3 \textsubscript{(1.9)}  & 2.26 \textsubscript{(0.37)} & 5346 & 0.16 & 376.3\\
				& \scalebox{1.15}{\ding{51}} & & 74.7 \textsubscript{(1.9)} & 2.24 \textsubscript{(0.37)} & 5626 & 0.18 & 396.4\\
				& &	\scalebox{1.15}{\ding{51}} & \textbf{75.4} \textsubscript{(1.9)} & \textbf{2.20} \textsubscript{(0.35)} & 14586 & 0.49 & 1103.8\\			\Xhline{1pt}
			\end{tabularx}
		}
		\label{tab:ab:refine}
		\vspace{-5pt}
	\end{table}
	\renewcommand{\arraystretch}{1}

	\vspace{1pt}
	\subsubsection{Recurrent Updater}
	We ablate the design of the recurrent updater from three aspects: feature integration, update components, and parameter sharing, as shown in Table \ref{tab:update}.
	First, we compare two strategies for feature integration: direct concatenation {\itshape vs.} explicit decoupling into motion-related and texture branches. The decoupled design yields better performance (74.7\% {\itshape vs.} 73.6\% Dice), suggesting improved alignment by reducing semantic interference.
	Second, we test different updater architectures. Replacing GRU with plain convolution reduces performance, highlighting the benefit of temporal memory. LSTM performs slightly worse than GRU, possibly attributed to the coarse-to-fine nature of our registration strategy, where large deformation trends are captured in early stages. In later stages, the memory unit primarily serves to refine local updates, and thus does not require the more complex gating mechanisms of LSTM. The GRU with simpler structure proves sufficient for propagating deformation cues across iterations.
	Lastly, we evaluate parameter sharing. Sharing parameters across all scales leads to degraded performance, while within-scale sharing offers a better balance of accuracy and efficiency. Our final model adopts feature decoupling, GRU, and scale-specific sharing.
	
	\vspace{1pt}
	\subsubsection{Multi-scale for Iteration}
	We evaluate the impact of iterative refinement across multiple scales: \scalebox{1.25}{\sfrac{1}{16}, \sfrac{1}{8}, \sfrac{1}{4}, \sfrac{1}{2}}. As shown in Table \ref{tab:ab:iteration}, performing iteration only at the \scalebox{1.25}{\sfrac{1}{8}} scale leads to poor performance (Dice 59.9\%). Adding iterations at the \scalebox{1.25}{\sfrac{1}{4}} scale significantly improves the Dice score to 70.3\%, and incorporating \scalebox{1.25}{\sfrac{1}{16}} further boosts performance to 74.0\%. These results suggest that multi-scale refinement is essential for capturing deformations of different magnitudes. 
	
	We also study the effect of increasing iteration counts at each scale. Results show that more iterations improve performance, especially at higher-resolution scales like \scalebox{1.25}{\sfrac{1}{2}}. However, the gain from combining additional iterations is sublinear, suggesting that benefits are synergistic rather than additive. Based on the trade-off between accuracy and computational cost, we set the iteration numbers to {3,3,2,2} for ReCorr and {1,1,1,1} for ReCorr-S.
	
	\renewcommand{\arraystretch}{1.15}
	\begin{table}[!b]
		\vspace{-5pt}
		\centering
		\setlength{\tabcolsep}{7.5pt}
		\caption[c]{Ablation results for robustness evaluation under synthetic perturbations.}
		\vspace{0pt}
		\scalebox{1}{
			\begin{tabular}{clll}
				\Xhline{1pt}
				&  &  &  \\[-1.4ex]
				\multirow{4}{*}[2.6ex]{\textbf{Deformation Type}} & \multirow{4}{*}[2.6ex]{\textbf{Level}} & \multicolumn{2}{c}{\textbf{OASIS} (\textit{w/o Pre-affine)}} \\
				\cmidrule(lr){3-4}
				& & \multicolumn{1}{c}{Dice(\%) $\uparrow$} & \multicolumn{1}{c}{HD95 $\downarrow$} \\
				\Xhline{0.7pt}    
				\vspace{-7pt}   
				&  &  &  \\
				No Augmentation  &  & 74.7 (2.1)  & 2.94 (0.36) \\[0.1ex]
				\Xhline{0.7pt}
				\vspace{-8.5pt}
				&  &  &  \\
				\multirow{4}{*}[1.2ex]{SVF} 
				& $s$=2  & 74.2 (2.0)  & 3.03 (0.40) \\
				& $s$=4  & 74.5 (2.1)  & 2.96 (0.49) \\          
				& $s$=8 & 74.4 (1.9) & 2.95 (0.47)
				\\[0.1ex]
				\Xhline{0.7pt}
				\vspace{-8.5pt}
				&  &  &  \\
				
				\multirow{9}{*}[-0.4ex]{Affine}
				& Offset (-0.2) & 71.7 (1.8)  & 3.29 (0.47) \\
				& Offset (-0.1) & 74.5 (2.3)  & 2.98 (0.52) \\
				& Offset (0.1) & 74.1 (2.3)  & 3.20 (0.57) \\
				& Offset (0.2) & 72.1 (1.9)  & 3.18 (0.44) \\
				[-0.5ex]
				\cmidrule(lr){2-4}
				& Scale (-0.2) & 73.4 (2.0)  & 3.07 (0.49) \\
				& Scale (-0.1)  & 74.0 (2.0) & 3.12 (0.50) \\
				& Scale (0.1)  & 73.3 (2.3) & 3.27 (0.59) \\
				& Scale (0.2) & 72.3 (2.3)  & 3.38 (0.75) \\ [0.1ex]	
				\Xhline{1pt}
			\end{tabular}
		}
		\label{tab:aug}
		\vspace{2pt}
	\end{table}
	\renewcommand{\arraystretch}{1}
	
	\vspace{1pt}
	\subsubsection{Original Resolution Operations}
	
	As shown in Table \ref{tab:ab:refine}, we evaluate three strategies at the original resolution: no operation (Plain), feature-based refinement (Refine), and one iteration (Iter.(1)). Adding a refinement step improves performance from 74.3\% to 74.7\% Dice with minimal increase in memory and time. Performing one full-resolution iteration yields further improvement (75.4\% Dice, 2.20~mm HD95), but incurs a substantial increase in GPU usage (×2.6 memory) and FLOPs (376G $\rightarrow$ 1103G). Considering the accuracy-efficiency trade-off, we adopt feature refinement at full resolution while keeping recurrent iterations at lower scales in our final model.

	\subsubsection{Sequence Supervision Strategy}

	We evaluate two strategies for applying sequence loss during training: one that supervises all intermediate deformation fields across iterations (``Full Sequence'') and another that only supervises the final output at each scale (``Last of scale"). As shown in Table~\ref{tab:loss}, \tred{the ``Full Sequence" strategy yeilds} slightly better accuracy (Dice: 74.7\% {\itshape vs.} 74.5\%) and lower HD95, {Although it incurs higher GPU memory, \tred{the additonal supervision provides more consistent optimization signals across iterations. Thus, we adopt the ``Full Sequence'' strategy for better optimization guidance and accuracy.}
	
	We further evaluate the impact of $\gamma$, which balances the relative weight of earlier and later predictions in the sequence loss. As shown in Fig.~\ref{fig:ab:gamma}, a value of $\gamma=0.7$ achieves the best Dice score, indicating an effective trade-off between emphasizing final predictions and preserving intermediate supervision. Therefore, we set $\gamma=0.7$ as the default for all experiments.
	
	\renewcommand{\arraystretch}{1.1}
	\begin{table}[!t]
		\vspace{0pt}
		\centering
		\setlength{\tabcolsep}{4pt}
		\caption[c]{Ablation of sequence supervision strategy. We compare two designs for the sequence loss with exponentially increasing weights: using all intermediate deformation fields (``Full Sequence") or only the final output at each scale (``Last of scale"). Settings used in our final model are underlined. The best value is shown in bold.}
		\scalebox{1}{
			\begin{tabularx}{\linewidth}{cc|ccc@{}}
				
				\Xhline{1pt}
				\vspace{-2pt}
				&  &  &  &\\[-0.6ex]
				\multicolumn{2}{c|}{\textbf{Experiment}} & \multicolumn{2}{c}{\textbf{OASIS} \textit{(w/o Pre-affine)}} & Train GPU \\
				&  &  &  &\\[-2.1ex]
				\underline{Full Sequence} & Last of scale & Dice \textsubscript{(\%)}$\uparrow$ & HD95 \textsubscript{(mm)} $\downarrow$ & (MB) \\[0.8ex]
				\Xhline{0.7pt}    
				&  &  &   \\[-2ex]
				\scalebox{1.15}{\ding{51}} &  & \textbf{74.7} \textsubscript{(1.9)} & \textbf{2.24} \textsubscript{(0.37)} & 23544 \\
				& \scalebox{1.15}{\ding{51}} & 74.5 \textsubscript{(2.0)} & 2.25 \textsubscript{(0.38)} & 19266 \\
				\Xhline{1pt}
			\end{tabularx}
		}
		\label{tab:loss}
		\vspace{0pt}
	\end{table}
	\renewcommand{\arraystretch}{1}

	\begin{figure}[]
		\vspace{2pt}
		\centerline{\includegraphics[width=0.47\textwidth]{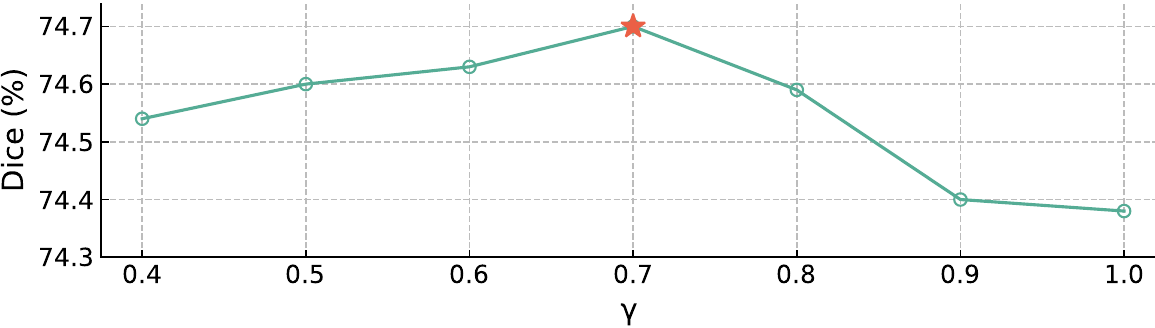}}
		\vspace{-3pt}
		\caption{Line graph of ablation results for the hyperparameter $\gamma$ of loss on the OASIS dataset without affine pre-registration. The orange star indicates the setting used in our final model.}
		\label{fig:ab:gamma}
		\vspace{-5pt}
	\end{figure}

	\subsubsection{Robustness Evaluation under Synthetic Perturbations}

	To evaluate the robustness of our model under varying deformation conditions, we conduct additional experiments on the non-affine OASIS dataset with synthetic perturbations. Specifically, we apply both non-linear and affine transformations to simulate diverse and challenging deformation scenarios. For non-linear perturbations, we introduce SVF-based fields at $s=2$, 4, and 8, corresponding to resolutions of (\scalebox{1.25}{\sfrac{1}{2}, \sfrac{1}{4}, \sfrac{1}{8}}), respectively. The $s=2$ setting introduces more local and fine-grained deformations, allowing us to assess the model's sensitivity to subtle spatial variations. For affine perturbations, we apply controlled global shifts (offsets of $\pm 0.1$ and $\pm 0.2$) and scaling factors along the x-axis.
	
	As shown in Table \ref{tab:aug}, the model maintains stable performance across most settings, with only moderate drops under strong perturbations. For SVF-based deformations, performance remains comparable to the baseline, confirming the robustness of the model to local non-linear distortions. In the affine setting, small offsets and scalings have minimal impact, while larger transformations (\textit{e.g.}, offset = $-0.2$ or scale = $0.2$) lead to a more noticeable decline in Dice, indicating the increased challenge posed by global shifts.
	
	\vspace{3pt}
	\section{Discussion \& Conclusion }
	\vspace{0pt}
	\label{sec:conclusion}
	
	This paper addresses the challenge of large deformation registration by proposing ReCorr, an efficient recurrent framework based on explicit voxel-to-region matching. While explicit feature matching strategies better align with the goal of voxel-level correspondence modeling, existing methods often suffer from high computational cost or limited search regions. ReCorr tackles this by performing iterative local search, where each step performs voxel-to-region matching within a small neighborhood and updates the search center accordingly.
	This enables the progressive convergence toward globally optimal alignment at minimal cost per iteration. To further reduce redundancy and improve alignment quality, ReCorr decouples motion-related and texture-related information into separate branches, allowing the network to focus on spatial correspondence without interference from irrelevant semantic cues. 
	Compared to prior works, ReCorr achieves superior or competitive accuracy across diverse experimental setups, while maintaining high efficiency in both computational cost and inference speed.

	However, several aspects remain to be explored. Adaptive strategies for the size of the search region could be used to improve the efficiency and accuracy of the network, such as using a larger search scope for the first iteration and a smaller one for the subsequent iterations. In addition, while local matching at low resolution is effective in many large deformation cases, it may struggle in extreme scenarios such as large-angle rotations or flipped structures, where local cues become ambiguous. In such cases, introducing sparse global matching may offer more reliable guidance and help improve convergence. Finally, adaptive attention mechanisms could help the network focus on informative regions rather than fixed local neighborhoods, improving alignment in ambiguous or low-contrast areas. These aspects could be further investigated to improve the performance of the network and its applicability in real-clinical scenarios.
	
	In summary, our work offers a robust and efficient solution for large deformation registration within the explicit matching paradigm, achieving a favorable balance between accuracy and computation.

	\vspace{2pt}
	\begin{small}
		\bibliographystyle{IEEEtran}
		\bibliography{References}
	\end{small}
	
\end{document}